\documentclass[10pt,twocolumn,letterpaper]{article}

\usepackage[pagenumbers]{cvpr} 

\usepackage{cuted} 

\usepackage{currfile} 

\usepackage{caption} 

\usepackage{microtype}

\usepackage{xspace}
\usepackage{multirow}
\usepackage{tabularx}
\usepackage[breakable,skins]{tcolorbox}
\tcbset{sbprompt/.style={breakable,enhanced,colback=cvprblue!3,colframe=cvprblue!55,boxrule=0.5pt,arc=1.5pt,left=4pt,right=4pt,top=3pt,bottom=3pt,fonttitle=\footnotesize\bfseries,coltitle=white,colbacktitle=cvprblue!75,fontupper=\footnotesize\raggedright,before upper={\setlength{\parskip}{0.35em}}}}
\newcommand{\sbench}{StudioBench\xspace}

\usepackage{tikz}
\usetikzlibrary{positioning,arrows.meta,fit,calc,backgrounds,decorations.pathreplacing,shapes.geometric}

\definecolor{bpOurs}{RGB}{0,114,178}      
\definecolor{bpPrior}{RGB}{213,94,0}      
\definecolor{bpData}{RGB}{0,158,115}      
\definecolor{bpAccent}{RGB}{230,159,0}    
\definecolor{bpGray}{RGB}{110,110,110}    
\definecolor{bpLight}{RGB}{238,238,238}   
\colorlet{bpOursFill}{bpOurs!18}
\colorlet{bpPriorFill}{bpPrior!15}
\colorlet{bpDataFill}{bpData!15}

\tikzset{
  bp/base/.style   ={draw=black!70, line width=0.5pt, rounded corners=2pt, align=center,
                     font=\sffamily\scriptsize, minimum height=0.6cm, inner sep=3pt},
  bp/module/.style ={bp/base, fill=white},
  bp/frozen/.style ={bp/base, fill=bpLight, draw=bpGray, text=black!80},
  bp/ours/.style   ={bp/base, fill=bpOursFill, draw=bpOurs, line width=0.7pt},
  bp/prior/.style  ={bp/base, fill=bpPriorFill, draw=bpPrior},
  bp/data/.style   ={bp/base, fill=bpDataFill, draw=bpData},
  bp/todo/.style   ={bp/base, dashed, draw=bpGray, text=bpGray},
  bp/flow/.style   ={-{Latex[length=1.6mm]}, line width=0.6pt, black!75},
  bp/note/.style   ={font=\sffamily\tiny, text=bpGray, align=center},
  bp/panel/.style  ={font=\sffamily\footnotesize\bfseries, anchor=west},
  bp/frame/.style  ={draw=black!70, fill=white, line width=0.4pt},
}

\AtEndPreamble{%
  \crefname{figure}{Figure}{Figures}%
  \Crefname{figure}{Figure}{Figures}%
  \crefname{table}{Table}{Tables}%
  \Crefname{table}{Table}{Tables}%
}

\definecolor{cinavy}{HTML}{156082}\definecolor{cired}{HTML}{B14E3B}
\newcommand{\cibaraxis}{\begin{tikzpicture}[x=0.22cm,baseline=-0.6ex]
  \draw[gray!70,line width=0.3pt] (-4,0) -- (5,0); \draw[gray!60,dashed,line width=0.3pt] (0,-0.13) -- (0,0.13);
  \node[font=\tiny,inner sep=0,anchor=south] at (-4,0.05) {$-4$}; \node[font=\tiny,inner sep=0,anchor=south] at (0,0.1) {$0$};
  \node[font=\tiny,inner sep=0,anchor=south] at (5,0.05) {$+5$};\end{tikzpicture}}
\newcommand{\cibarbase}[4]{\begin{tikzpicture}[x=0.22cm,baseline=-0.6ex]
  \draw[gray!45,line width=0.25pt] (-4,0) -- (5,0); \draw[gray!60,dashed,line width=0.3pt] (0,-0.13) -- (0,0.13);
  \draw[line width=0.9pt,#4] (#2,0) -- (#3,0); #1\end{tikzpicture}}
\newcommand{\cibar}[4]{%
  \ifstrequal{#4}{axis}{\cibaraxis}{%
  \ifstrequal{#4}{up}{\cibarbase{\fill[cinavy] (#1,0) circle (0.08);}{#2}{#3}{cinavy}}{%
  \ifstrequal{#4}{down}{\cibarbase{\fill[cired] (#1,0) circle (0.08);}{#2}{#3}{cired}}{%
  \cibarbase{\fill[white] (#1,0) circle (0.07); \draw[gray,line width=0.5pt] (#1,0) circle (0.07);}{#2}{#3}{gray}}}}}
\makeatletter
\renewcommand\paragraph{\@startsection{paragraph}{4}{\z@}%
  {1.2ex \@plus 0.4ex \@minus 0.2ex}{-1em}{\normalfont\normalsize\bfseries}}
\makeatother

\definecolor{cvprblue}{rgb}{0.21,0.49,0.74}
\usepackage[breaklinks,colorlinks,allcolors=cvprblue]{hyperref}
\definecolor{main}{HTML}{4472C4} 
\definecolor{sub}{HTML}{EBF4FF}
\newtcolorbox{boxA}{
  enhanced, breakable,
  boxrule = 0pt,
  colback = sub,
  top = 3pt, bottom = 3pt, left = 5pt, right = 5pt,
  before skip = 5pt, after skip = 5pt,
  borderline west = {2pt}{0pt}{main}, 
  borderline east = {2pt}{0pt}{main}, 
}

\newcommand{\github}{\raisebox{-1.5pt}{\includegraphics[height=1.0em]{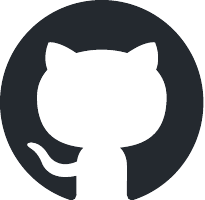}}\xspace}
\newcommand{\web}{\raisebox{-1.5pt}{\includegraphics[height=1.0em]{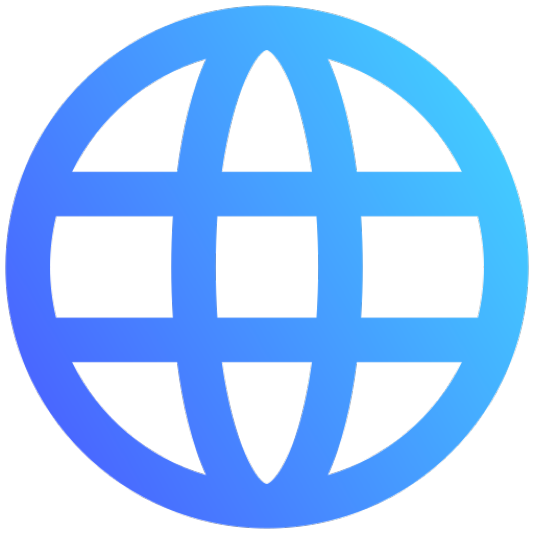}}\xspace}

\newcommand{\studiologo}{\raisebox{-1.5pt}{\includegraphics[height=1.0em]{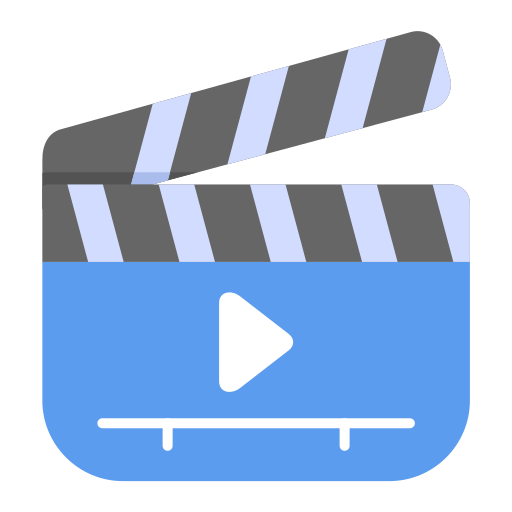}}\xspace}
\newcommand{\canvas}{\raisebox{-1.5pt}{\includegraphics[height=1.0em]{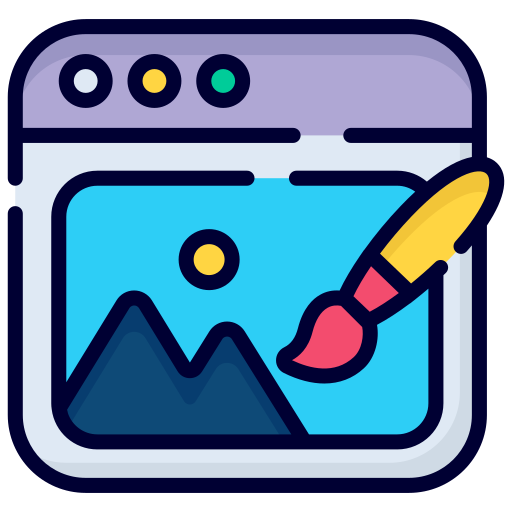}}\xspace}

\def\confName{CVPR}
\def\confYear{2027}

\title{StudioBench: Benchmarking Agentic Multimodal Large Language Models for Productive Audiovisual Creation}

\author{Xinglong Xu$^{1}$\thanks{Equal contribution}, Yuxia Zhu$^{2*}$, Siyuan Li$^{3}$, Lixu Tian$^{4}$, Xiaofei Zhao$^{4}$, Yuxin Liu$^{3}$, Caijun Jia$^{3}$,\\ Jintao Chen$^{5}$, Bihui Yu$^{1}$, Alden Cai$^{3}$, Asher Qin$^{3\dagger}$, Jingxuan Wei$^{4\dagger}$, Cheng Tan$^{3}$\thanks{Corresponding author}
\\
$^1$University of Chinese Academy of Sciences 
$^2$Peking University $^3$Tencent \\$^4$Qilu University of Technology $^5$Zhejiang University
}

\begin{document}
\maketitle

\begin{abstract}
Video generators can now render convincing short clips, yet complete audiovisual works require coordinating visuals, sound, and editing across multiple stages. Existing benchmarks focus on individual generators or video agents under narrower production settings, leaving end-to-end creation by general multimodal large language models (MLLMs) underexplored.
We introduce \sbench, a benchmark comprising 160 tasks across eight video genres and five task types, spanning diverse audiovisual production requirements. Models produce minutes of finished audiovisual content using 33 atomic generation, editing, and inspection tools, without shell access or workflow templates. We evaluate the delivered films, while analyzing reconstructed workflow graphs to characterize production behavior and diagnose failures.
Across seven MLLMs, explicit requirements are nearly saturated, yet production quality varies substantially in picture clarity and composition, object structure, motion, and lip sync. All models spontaneously organize multi-stage production workflows without predefined templates. However, higher-quality models exhibit more selective production workflows, whereas additional tool calls alone show little association with quality. These findings identify the missing quality loop of effective inspection, selection, and revision as a key bottleneck in productive audiovisual creation.
\par\vspace{0.5em}
\noindent\makebox[\linewidth][c]{%
\github~\href{https://github.com/openraiser/StudioBench}{\textbf{Code}}\quad 
\web~\href{https://openraiser.github.io/StudioBench/}{\textbf{Website}}\quad
\canvas~\href{https://openraiser.github.io/StudioBench/_canvas/index.html/}{\textbf{Canvas}}\quad
\studiologo~\href{https://github.com/openraiser/StudioFlow}{\textbf{StudioFlow}}
}
\par
\end{abstract}
\vspace{-0.8em}

\section{Introduction}
\label{sec:intro}

Modern text-to-video models can generate convincing short clips from a single prompt~\cite{openai2024sora,polyak2024moviegen,kong2024hunyuanvideo,wan2025wan,gao2025seedance}.
However, producing complete audiovisual works requires much more than generating individual clips. It involves maintaining visual and narrative consistency across scenes, coordinating speech, music, and sound effects with visual events, and assembling these elements into a coherent final product.
While specialized tools can generate individual components, their coordination requires planning, inspection, and revision.
The central challenge thus extends beyond generating high-quality clips to autonomously managing the entire audiovisual production process.

\begin{figure*}
\centering
\includegraphics[width=\linewidth]{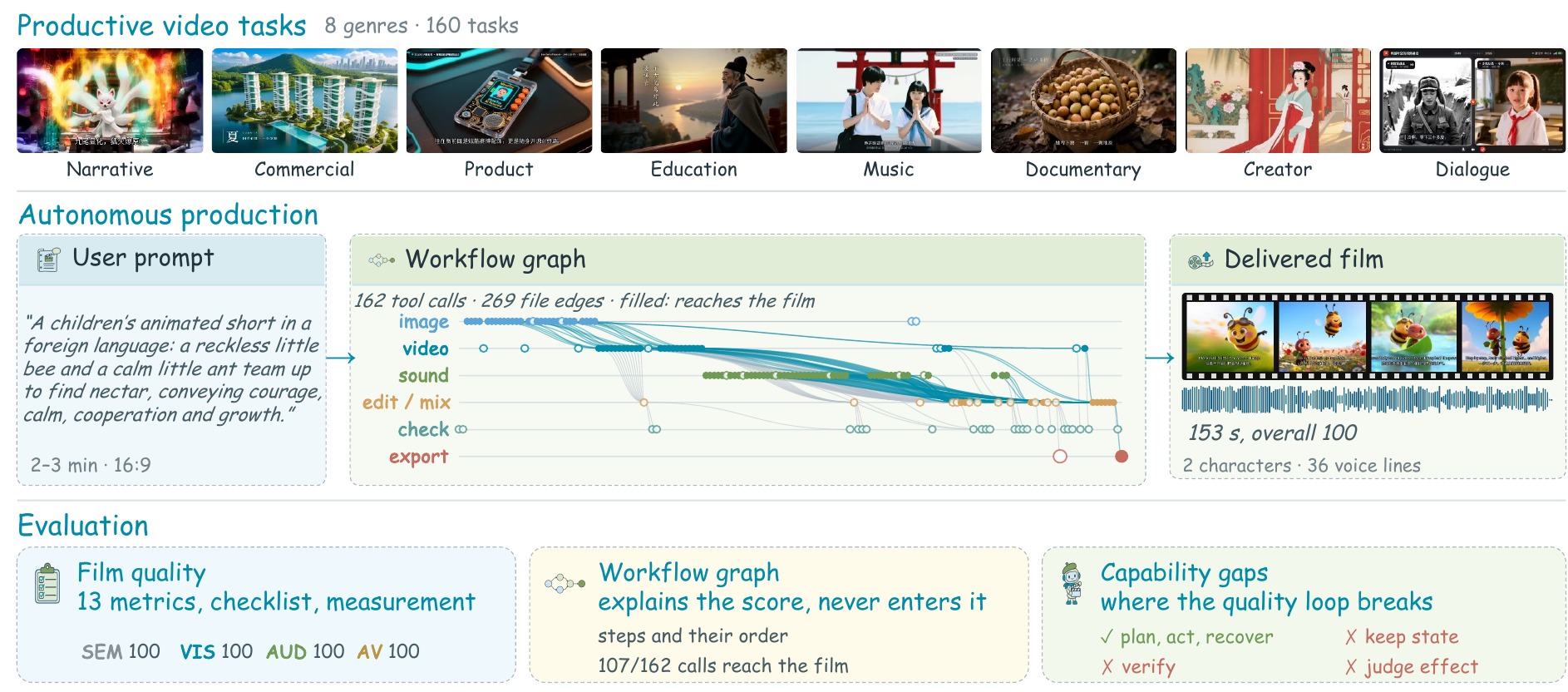}
\captionsetup{hypcap=false}
\captionof{figure}{\textbf{Overview of StudioBench.} The benchmark spans 160 tasks across eight video genres. Given a production brief and 33 atomic tools, an MLLM autonomously constructs a multi-stage workflow and delivers a complete audiovisual work. The film is scored independently, while its workflow graph supports behavioral analysis and capability diagnosis.}
\captionsetup{hypcap=true}
\label{fig:teaser}
\end{figure*}

Evaluation has largely followed generation at the level of individual generators rather than the agentic capabilities required for end-to-end production. VBench and VBench-2.0~\cite{huang2024vbench,zheng2025vbench2} evaluate video quality and semantic fidelity, while recent audiovisual benchmarks~\cite{hua2025vabench,liu2026javisdit,liu2026longavcompass} extend evaluation to sound and longer sequences. These benchmarks assess generated outputs rather than the multi-stage decisions required to produce complete audiovisual works. Benchmarks of video agents move closer to end-to-end production, but typically evaluate systems built around predefined workflows or composition skills~\cite{chen2026directorbench,wei2026videoweaver}, or rely on code-based rendering instead of generative video tools~\cite{diao2026vibevideobench}. Consequently, the agentic capabilities of general MLLMs remain insufficiently understood: whether they can independently organize and execute multi-stage audiovisual workflows using atomic tools, and how their inspection, selection, and revision decisions influence the quality of the final work.

We introduce \sbench, an end-to-end benchmark for evaluating the agentic capabilities of general MLLMs in productive audiovisual creation, as illustrated in Figure~\ref{fig:teaser}. By productive, we refer to audiovisual works created for real audiences, such as advertisements, educational videos, and music performances, rather than isolated demonstration clips. The benchmark comprises 160 tasks across eight genres and five task types. Given a production brief and 33 atomic tools spanning visual and audio generation, editing, inspection, and delivery, each model must autonomously plan, generate, inspect, revise, and assemble a complete audiovisual work without shell access or predefined workflow templates. We evaluate the delivered film using 13 whole-film metrics covering semantics, visual quality, audio quality, and audiovisual alignment, supplemented by requirement checklists and objective measurements of production defects. Separately, we reconstruct a \emph{workflow graph} from each execution trace, where nodes represent tool calls and edges capture artifact dependencies. By relating production decisions to final film quality, we can distinguish the ability to execute a complete workflow from the ability to effectively evaluate and improve its outputs. The entire execution, evaluation, and trace analysis pipeline is automated, requiring no human intervention during benchmark runs.

We evaluate seven MLLMs across all 160 tasks and analyze their performance at three levels.
At the product level, explicit requirements are nearly saturated, with the mean film satisfying 98.1\% of its requirement items, yet substantial quality differences remain in picture clarity and composition, body and object structure, motion, and lip sync.
At the process level, all models spontaneously organize multi-stage production workflows without predefined templates, commonly generating keyframes before video, inspecting intermediate outputs, and mixing audio before export.
At the capability level, higher-quality models exhibit more selective production workflows, while additional tool calls alone show little association with quality.
Moreover, models recover from most of reported tool errors but frequently fail to detect or correct perceptual defects that raise no explicit error.

These findings highlight the potential of general MLLMs for autonomous audiovisual production, while revealing persistent challenges in maintaining quality across stages. Closing the quality loop through effective inspection, selection, and revision remains a key direction for improvement. Our contributions are summarized as follows:
\begin{itemize}
\item \textbf{StudioBench}, a benchmark of 160 audiovisual creation tasks, spanning eight genres and five task types with fine-grained requirement checklists.
\item \textbf{An automated evaluation framework} with 33 atomic tools, enabling end-to-end MLLM execution, film-quality evaluation, and workflow-based behavioral analysis.
\item \textbf{A systematic empirical analysis}, revealing emergent production workflows, persistent quality gaps, and the missing quality loop as a key bottleneck in audiovisual creation.
\end{itemize}

\section{Related Work}
\label{sec:related}

\begin{table*}[t]
\centering
\caption{\textbf{Comparison with existing video and audiovisual benchmarks.} StudioBench evaluates general MLLMs on end-to-end audiovisual production. Unlike generator-level benchmarks, it assesses complete delivered works and analyzes their production processes.}
\label{tab:related}
\small
\setlength{\tabcolsep}{0pt}
\begin{tabular*}{\linewidth}{@{\extracolsep{\fill}}llclll@{}}
\toprule
Work & Evaluates & Length & Production Interface & Audio & Workflow \\
\midrule
VBench \cite{huang2024vbench}
& Generator & $\sim$2\,s & Generator & $\times$ & -- \\
VBench-2.0 \cite{zheng2025vbench2}
& Generator & 5--10\,s$^*$ & Generator & $\times$ & -- \\
LongAV-Compass \cite{liu2026longavcompass}
& Generator & 1--2\,min & Generator & Generated audio & -- \\
\midrule
VideoGen-Agent \cite{li2026videogenagent}
& Agent & $\leq$15\,s & Generator & $\times$ & $\checkmark$ \\
VideoWeaver \cite{wei2026videoweaver}
& Agent & $\sim$1\,min & Generator + Skills & Speech & $\checkmark$  \\
VibeVideoBench \cite{diao2026vibevideobench}
& Agent & 8\,s--15\,min & Code & Speech + code$^\dagger$ & $\times$ \\
\midrule
{\sbench}
& {Agent} & {1--3\,min}
& {Generator + Code + Atomic tools} & {Speech/music/SFX}
& $\times$ \\
\bottomrule
\end{tabular*}
\vspace{1mm}
\parbox{\linewidth}{\footnotesize
$^*$ Model-dependent.
$^\dagger$ Speech uses TTS; music and sound effects are synthesized in code.
$\times$ indicates absence; -- indicates not applicable.
Workflow Support refers to production guidance or reusable procedural skills supplied to the agent, rather than workflows constructed autonomously.
}
\end{table*}

\paragraph{Video generation benchmarks.}
Benchmarks for video generation primarily evaluate outputs rendered from prompts or references. VBench~\cite{huang2024vbench,huang2025vbenchpp} assesses 16 dimensions of video quality and prompt consistency, while VBench-2.0~\cite{zheng2025vbench2} extends evaluation to intrinsic faithfulness, including physics and commonsense. Audio-video benchmarks~\cite{hua2025vabench,liu2026javisdit} incorporate sound, with recent extensions to multi-shot sequences~\cite{wei2026msavbench} and longer videos~\cite{liu2026longavcompass}. FilmBench~\cite{wang2026filmbench} draws prompts from award-winning films. However, these benchmarks do not evaluate the multi-stage decisions required to generate, inspect, and assemble complete audiovisual works. In contrast, \sbench evaluates these agentic capabilities, using whole-film metrics rather than focusing primarily on clip-level quality.

\paragraph{Video production agents.}
Multi-scene and long-video systems typically use language models to plan scripts or storyboards before generating videos shot by shot \cite{lin2023videodirectorgpt,long2024videostudio,xie2024dreamfactory,li2024animdirector,wu2025movieagent,huang2026vimax}. Some incorporate narration and music \cite{xu2025mmstoryagent,wang2026mavis}, synchronize edits with music \cite{tang2025automv}, or render plans in 3D engines \cite{xu2025filmagent}. 
However, their workflows are predefined and evaluated as complete systems on their own prompts. VideoGen-Agent \cite{li2026videogenagent} trains a multimodal agent via reinforcement learning to coordinate generation and verification tools under task-specific tool-use patterns, evaluating on 600 LLM-written prompts for silent videos up to 15 seconds. LIT \cite{jian2026lit} co-evolves a director agent with a simulated task-generating user and a meta agent that refines its directing strategy.

\paragraph{Benchmarks of video agents.}
Recent benchmarks evaluate video-making or editing agents, often with predefined workflows or constrained production tools.
DirectorBench~\cite{chen2026directorbench} compares language models across four fixed workflows, DramaChain Bench~\cite{shi2026dramachain} evaluates stages of a short-drama pipeline, and Determine~\cite{verma2026determine} assesses operations in a predefined production graph.
AgenticVBench~\cite{cao2026agenticvbench} evaluates agents editing existing footage rather than creating videos from briefs.
VideoWeaver~\cite{wei2026videoweaver} uses learned composition skills to produce minute-long videos, with audio generation limited to speech.
VibeVideoBench~\cite{diao2026vibevideobench} evaluates video creation with sound, but relies on code-based rendering.
\sbench evaluates general MLLMs on minutes of audiovisual works, as shown in \cref{tab:related}.
General agentic MLLMs orchestrate image, video, and audio generators through atomic tools without predefined workflows.

\paragraph{Agent benchmarks.}
Tool-use, web, computer-use, and software-engineering benchmarks
\cite{qin2024toolllm,liu2024agentbench,mialon2024gaia,zhou2024webarena,xie2024osworld,jimenez2024swebench,yao2024taubench,merrill2026terminalbench,li2026toolathlon}
evaluate multi-step planning and execution, typically through automatic checks of outputs or environment states.
Benchmarks of professional deliverables
\cite{patwardhan2026gdpval,mazeika2025rli,starace2025paperbench}
and agent-generated artifacts \cite{zhu2025paper2video} instead assess final products using expert judgments, rubrics, or agent judges \cite{zhuge2025agentjudge}.
Audiovisual works require perceptual evaluation, as defects may produce no explicit errors.
We therefore employ a multimodal judge, following common practices
\cite{zheng2023mtbench,he2024videoscore,han2025videobench,cao2026t2avcompass}.

Language models can correct errors with explicit feedback \cite{gou2024critic}, but struggle without it \cite{huang2024cannotselfcorrect,kamoi2024selfcorrection}. Tool-using agents similarly detect reported errors more reliably than plausible but incorrect outputs \cite{kraishan2026loudquiet}, while visual inspection by code-based video agents does not reliably predict quality \cite{diao2026vibevideobench}. \sbench examines these perceptual verification failures in audiovisual production (\cref{sec:rq3}). Existing process evaluations assess agent trajectories \cite{ma2024agentboard} or compare workflows against reference graphs \cite{qiao2025worfbench}. Unlike Determine's predefined production graph, our workflow graphs are reconstructed from executed artifact dependencies and used for attribution rather than scoring.

\section{Preliminaries}
\label{sec:prelim}

\subsection{Task Formulation}
\label{sec:task}

We study whether a MLLM, given no workflow prior, can carry a productive film from its brief to delivery.
Each task gives a brief $w$, a duration window $[a, b]$, and an aspect ratio $\rho$.
Each task is also labelled with the sound types it depends on, $A \subseteq \{\textit{speech}, \textit{music}, \textit{sfx}, \textit{dialogue}\}$.

\vspace{-1mm}
\paragraph{Requirement checklist.}
From $w$ we split a checklist $R = \{r_1, \dots, r_k\}$ of atomic requirements, each of which a judge can mark as met, partly met, not met or not applicable.
$R$ is used only for scoring and never shown to the model.

\vspace{-1mm}
\paragraph{Execution and delivery.}
The model under test acts as a policy $\pi$ over a tool set $\mathcal{T}$ and produces a trajectory $\tau = (c_1, o_1, \dots, c_n, o_n)$ of tool calls $c_i$ and observations $o_i$.
A run ends when the model delivers a film $v$ through the \texttt{export} tool; a run without an export counts as not delivered.
Each run has a budget of 200 tool calls and a wall-clock limit of 4 hours for tasks of 1--2 minutes and 8 hours for tasks of 2--3 minutes.

\vspace{-1mm}
\paragraph{Workflow graph.}
A trajectory lists calls in time order, but production is a dependency structure: a keyframe feeds a clip, the clip feeds a trim, the trim feeds the edit.
We therefore represent each run by its workflow graph $G(\tau) = (V, E)$, in which every tool call $c_i$ is a node and an edge $c_i \rightarrow c_j$ means that $c_j$ read a file written by $c_i$.
The environment records the producer of every file a call reads, so $G$ is recovered exactly from the logs.
The ancestors of the final \texttt{export} node are the part of the work that reaches the film; a check whose output feeds no later production node changed nothing in it.

\vspace{-1mm}
\paragraph{Unit of evaluation.}
The unit is one run: a delivered film $v$ together with its workflow graph $G(\tau)$.
The film supports the product analysis (RQ1, \cref{sec:rq1}), the graph supports the process analysis (RQ2, \cref{sec:rq2}), and the two together support the capability analysis (RQ3, \cref{sec:rq3}).

\subsection{Atomic Tool Environment}
\label{sec:env}

\begin{figure}[t]
\centering
\includegraphics[width=\linewidth]{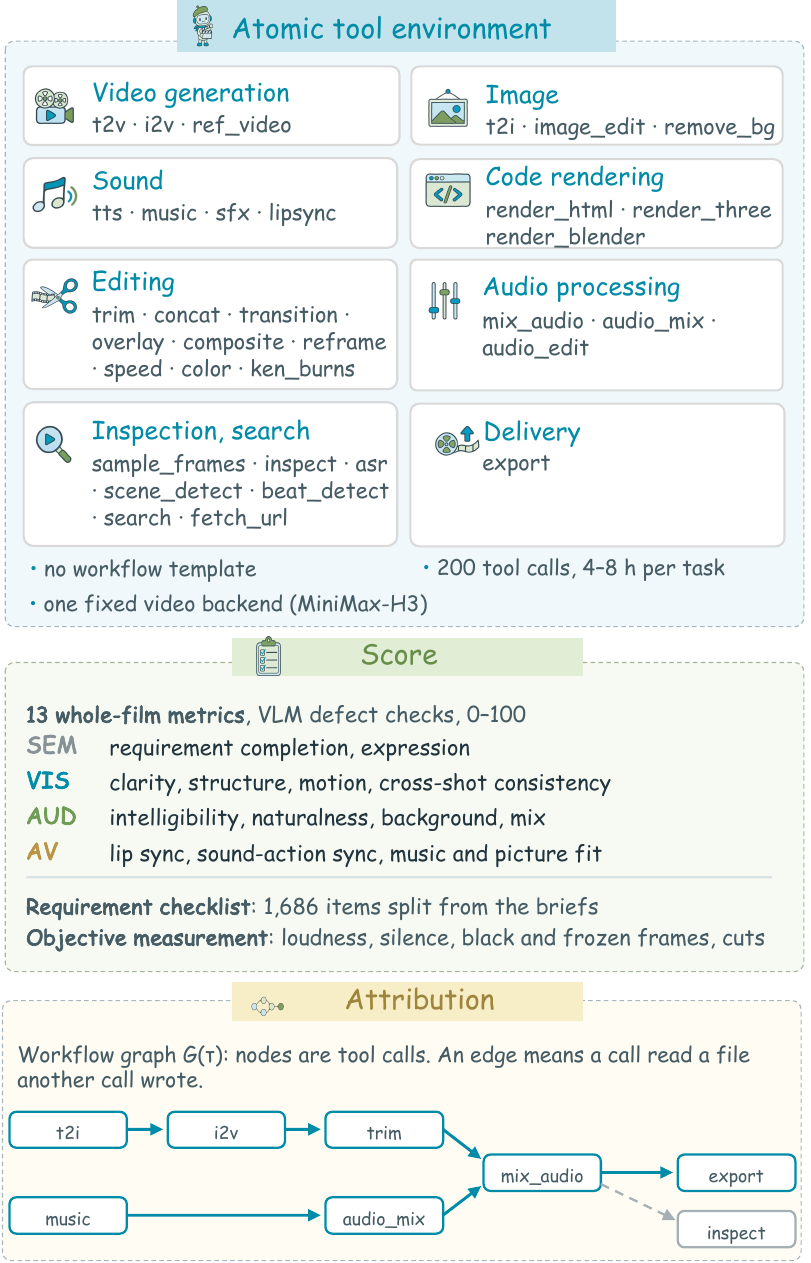}
\caption{\textbf{Atomic tools in; the film is scored, the workflow graph explains.}
Top: the 33 tools in eight groups. Bottom: metrics, checklist and measurement form the score, while the workflow graph $G(\tau)$ only attributes it.}
\label{fig:env}
\end{figure}

The model under test sees only 33 atomic tools in eight groups (\cref{fig:env}): video generation, image, sound, code rendering, editing, audio processing, inspection and search, and delivery.
All video generation goes through one backend, MiniMax-H3 \cite{minimax2026h3}, which produces 5\,s or 8\,s segments with their own audio track.
Fixing the generator removes differences in rendering quality between runs, so that the remaining variance comes from how the model directs.

The environment is deliberately minimal.
There is no shell, and code can only be written into the arguments of the rendering tools.
The system prompt describes what each tool does and the budget, and contains no workflow template.
Each task runs in its own working directory, internet access goes only through the search tools, and the source video of the task is blocked from search.
Each call is logged with its arguments, input and output files, errors and duration. 

\section{\sbench}
\label{sec:benchmark}

\subsection{Task Construction}
\label{sec:construction}

\noindent\textbf{Source and screening.}
\sbench is built by rephrasing publicly released AIGC videos. We collected AIGC video works from public platforms and, after manual review, kept 1{,}276 candidate tasks.
Each candidate consists of a one-sentence brief, a structured full description of the work for reference, and labels for genre and task type.

\noindent\textbf{Genres and task types.}
\sbench covers eight genres, each chosen for the production ability it stresses: \textit{Narrative} (long-range storytelling), \textit{Commercial} (an advertisement under constraints), \textit{Product} (faithful product information), \textit{Education} (explaining knowledge), \textit{Music} (music as the timeline), \textit{Documentary} (organizing information), \textit{Creator} (online-native formats) and \textit{Dialogue} (speech and lip sync).
Tasks fall into five types: \textit{Open Brief}, \textit{Constraint}, \textit{Asset-conditioned}, \textit{Reference-conditioned} and \textit{Audio-led}.

\noindent\textbf{Sampling and statistics.}
We sampled 20 tasks per genre, balanced across task types within each genre where the pool allows, for 160 tasks in total (\cref{fig:dataset} in the appendix).
Target durations range from 60 to 180 seconds; 129 tasks are 16:9 landscape and the rest are portrait or wide-screen.
Sound labels are multi-label: 157 tasks need music, 128 sound effects, 89 speech and 55 dialogue.

\noindent\textbf{Requirement checklists.}
We split each brief into atomic requirements, 1{,}686 in total, with 4--18 per task and a median of 10.
Of these, 545 are marked critical, such as the duration, the aspect ratio, a named landmark, a specified caption or ``no voice-over''.
By content, 787 requirements concern what is on screen, 321 the format, 264 the sound, 127 style, 117 structure and order, 54 consistency and 16 on-screen text.

\begin{figure*}[t]
\centering
\includegraphics[width=\linewidth]{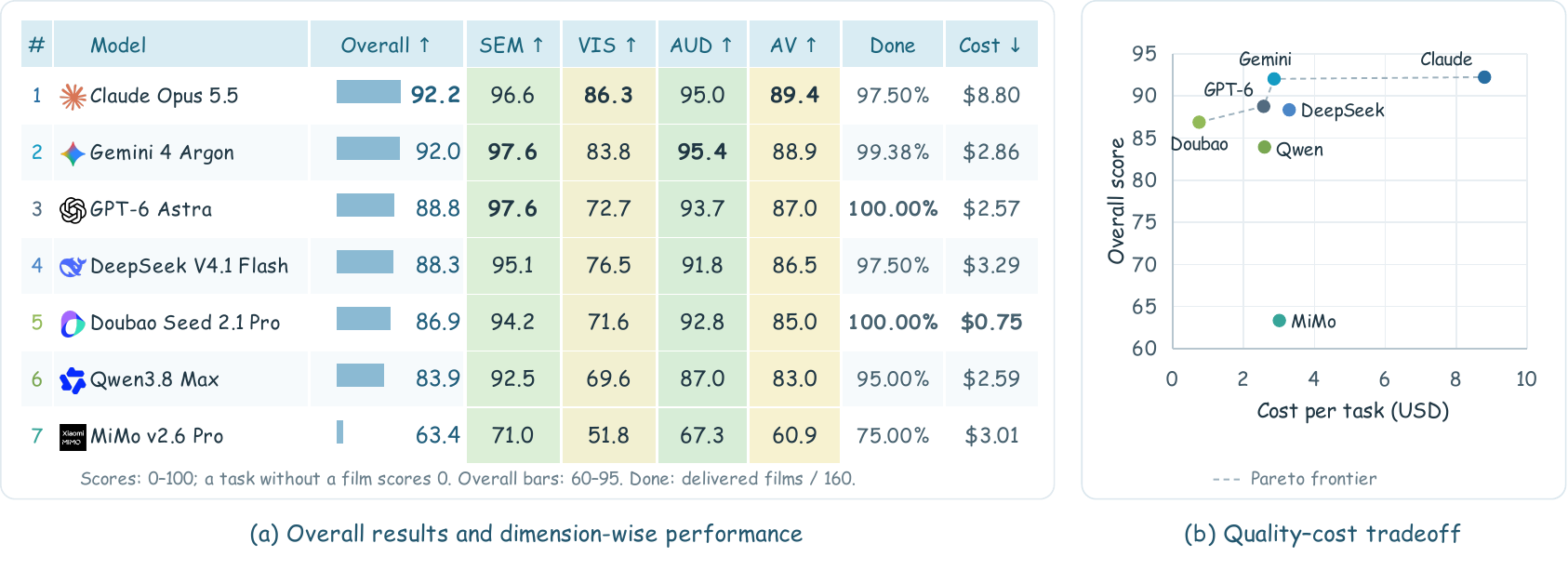}
\caption{\textbf{Product quality and cost comparison across seven MLLMs.}
(a) Overall rankings and performance across four dimensions: semantics (SEM), visual quality (VIS), audio quality (AUD), and audiovisual alignment (AV), averaged over all 160 tasks with 0 for a task without a film. Done indicates delivered films out of 160.
(b) Quality--cost trade-off against the median LLM cost per task, with the dashed line indicating the Pareto frontier. Overall scores range from 0 to 100.}
\label{fig:product_quality}
\end{figure*}

\subsection{Scoring Protocol}
\label{sec:scoring}



\paragraph{Whole-film metrics.}
A VLM judge \cite{han2025videobench,cao2026t2avcompass} inspects each film across six modality-separated calls and answers 73 time-stamped defect checks rather than assigning grades, assisted by candidate segments from programmatic checks of loudness~\cite{itu2023bs1770,ebu2023r128}, silence, clipping, blur, black or frozen frames, shake, and mouth motion.
Averaged defect costs over two passes (and any fatal check) map via calibrated thresholds (\cref{tab:cuts}) to three levels $\{0, 0.5, 1\}$ on 12 craft metrics, joining requirement completion (met $1$, partly met $0.5$, not met $0$; inapplicable items excluded) to form 13 metrics in four dimensions (\cref{fig:env,tab:metrics}).
Semantics weights requirement completion and expression 2:1, and the overall score weights dimensions by genre (\cref{tab:cuts}), scoring undelivered runs as $0$ (Appendix~\ref{sec:supp_metrics}).

We additionally report objective file defects (\cref{tab:defects}); workflow graphs are analyzed separately for behavioral diagnosis and failure attribution, never entering the score.

\section{Experiments}
\label{sec:exp}

We systematically evaluate the agentic audiovisual production capabilities of general MLLMs through three questions:

\vspace{2mm}
\noindent\textbf{RQ1: Product Quality.}
How well do MLLMs satisfy production requirements, and how do their audiovisual quality and cost compare?

\vspace{2mm}
\noindent\textbf{RQ2: Workflow Emergence.}
Can MLLMs autonomously organize multi-stage production workflows without predefined templates, and how do their strategies differ? 

\vspace{2mm}
\noindent\textbf{RQ3: Capability Bottlenecks.}
Which agentic capabilities enable successful production, and what limitations prevent models from reliably improving the final quality? 

\subsection{Experimental Setup}
\label{sec:setup}

We evaluate seven representative MLLMs: Claude Opus 5.5~\cite{anthropic2026opus55}, Gemini 4 Argon~\cite{google2026gemini4}, DeepSeek V4.1 Flash~\cite{deepseek2026v41}, Qwen3.8 Max~\cite{qwen2026qwen38}, GPT-6 Astra~\cite{openai2026gpt6}, MiMo v2.6 Pro~\cite{xiaomi2026mimo26}, and Doubao Seed 2.1 Pro~\cite{bytedance2026seed21}.
All seven models are evaluated on the same 160 tasks under identical system prompts, tool sets, generation backends, and budgets, with one run per task--model pair, yielding 1{,}120 runs in total.
Of these, 1{,}063 deliver a film, and every delivered film is scored.
A run without a film scores 0, so each model is averaged over all 160 tasks, and 95\% rank intervals come from 2{,}000 bootstrap resamples of the tasks.
The 57 undelivered runs and their causes are detailed in Appendix~\ref{sec:supp_infra}.
As a robustness check, excluding the 13 runs interrupted by the model API or the time limit raises Claude Opus 5.5 to 94.6, swaps DeepSeek V4.1 Flash with GPT-6 Astra and Qwen3.8 Max with Doubao Seed 2.1 Pro, while preserving the other positions. Removing the 150 delivered runs with five or more service-side tool errors only swaps the top two models.

\subsection{RQ1: Product Quality}
\label{sec:rq1}
\vspace{2mm}
\begin{boxA}
\noindent\textbf{Takeaway:}
Agentic MLLMs nearly saturate explicit requirements, while substantial differences in audiovisual craft distinguish their overall quality, with notable quality--cost trade-offs.
\end{boxA}
\vspace{2mm}
\noindent\textbf{Overall quality reveals four tiers.}
\cref{fig:product_quality}(a) presents the overall leaderboard and dimension-wise performance.
Claude Opus 5.5 (92.2) and Gemini 4 Argon (92.0) lead without a significant difference, followed by GPT-6 Astra (88.8), DeepSeek V4.1 Flash (88.3), and Doubao Seed 2.1 Pro (86.9), then Qwen3.8 Max (83.9) and MiMo v2.6 Pro (63.4).
MiMo v2.6 Pro trails because it delivers only 75\% of its tasks.

\begin{figure*}[t]
\centering
\includegraphics[width=\linewidth]{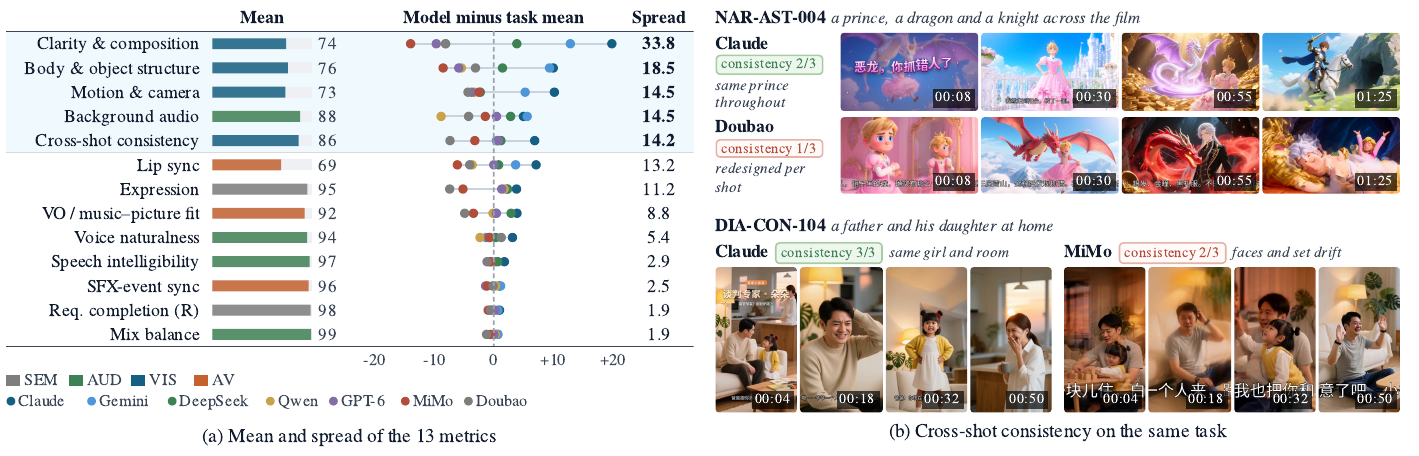}
\caption{\textbf{Requirements are saturated; craft separates the models.}
(a) Mean of each metric over delivered films (bars, by dimension), each model's deviation from the task mean (dots) and the range between models; the five shaded metrics differ by at least 14 points. Sound and audio-visual metrics leave out the six silent films, which receive the lowest level on them by rule (Appendix~\ref{sec:supp_metrics}).
(b) Films on the same task with the judge's cross-shot consistency level (1--3): Opus 5.5 keeps its characters, while Doubao and MiMo let them drift.}
\label{fig:where}
\end{figure*}

\noindent\textbf{Explicit requirements saturate, but craft differentiates models.}
The mean film satisfies 98.1\% of its requirement items, and 81\% of films satisfy every checklist item.
Expression and sound scores remain high, while visual quality varies substantially.
\cref{fig:where}(a) shows the largest inter-model gap in clarity and composition, followed by body and object structure, motion, background audio, cross-shot consistency, and lip sync.
\cref{fig:where}(b) illustrates these differences: Claude maintains character consistency across shots, whereas Doubao and MiMo exhibit noticeable identity drift on the same tasks.

\noindent\textbf{Quality exhibits notable cost trade-offs.} \cref{fig:product_quality}(b) compares quality against median LLM cost per task, with the dashed line marking the Pareto frontier. Gemini scores 92.0 at \$2.86, just 0.2 points below Claude at \$8.80, while Doubao achieves 86.9 at only \$0.75. The frontier spans Doubao, GPT-6, Gemini, and Claude, showing that higher expenditure does not yield proportional quality gains.


\subsection{RQ2: Workflow Emergence}
\label{sec:rq2}
\vspace{2mm}

\begin{boxA}
\noindent\textbf{Takeaway:}
Agentic MLLMs spontaneously organize complete multi-stage audiovisual workflows without predefined templates, while exhibiting distinct strategies in planning, asset selection, and revision.
\end{boxA}

\vspace{2mm}

\begin{figure}[t]
\centering
\includegraphics[width=\linewidth]{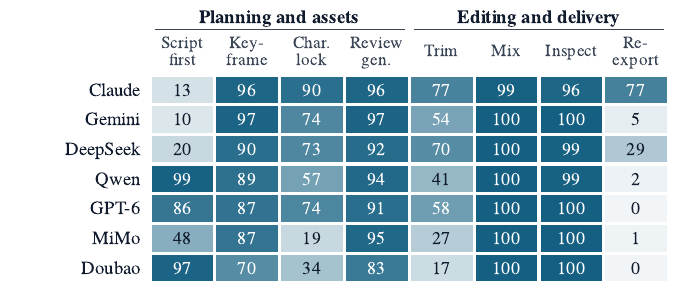}
\caption{\textbf{A production workflow emerges without a template.} Share of delivered runs (\%) whose workflow graph contains each step. Keyframes, review, mixing and inspection are near universal; scripting, character, trimming and re-export separate the models.}
\label{fig:workflow}
\end{figure}

\paragraph{A common production workflow emerges.} As shown in Figure~\ref{fig:workflow}, all MLLMs independently establish multi-stage production workflows. Keyframe generation precedes video generation in 70--97\% of runs, generated clips are reviewed in 83--97\%, audio is mixed before export in 99--100\%, and final films are inspected in 96--100\%. Despite receiving no templates, models consistently organize generation, inspection, editing, and delivery into coherent production pipelines.
\vspace{2mm}

\paragraph{Models adopt distinct strategies.} Doubao, Qwen3.8 Max, and GPT-6 Astra write full scripts before tool use in 86--99\% of runs, whereas Claude, Gemini, and DeepSeek do so in only 10--20\%, often planning during production. Character locking varies from 19\% to 90\% of runs, clip trimming from 17\% to 77\%, and re-exporting after review from 0\% to 77\%. These differences reveal substantial variation in how models organize and refine their productions.

\begin{figure*}[t]
\centering
\includegraphics[width=\linewidth]{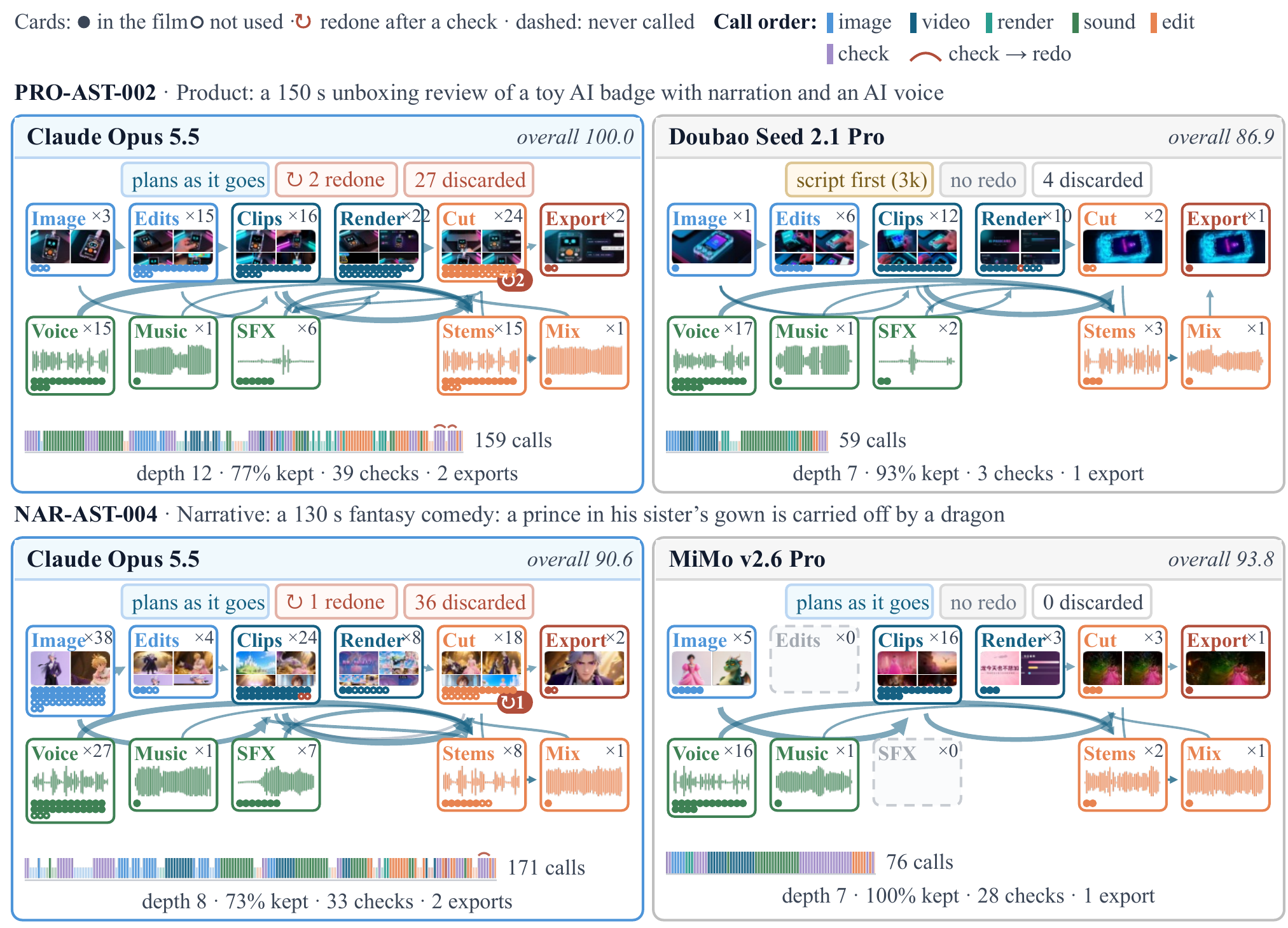}
\caption{\textbf{Emergent audiovisual production workflows.}
Colored nodes represent production tools, artifact dependencies trace the delivered film, and timelines show tool-call sequences. 
}
\label{fig:workflow_case}
\end{figure*}

\paragraph{Paired workflows illustrate these differences.}
Figure~\ref{fig:workflow_case} compares execution traces on identical tasks. In NAR-AST-004, Claude and MiMo both plan during production, but Claude makes 171 tool calls and discards 36 alternatives, retaining 73\% of production nodes, whereas MiMo makes 76 calls and retains all generated production nodes. These examples illustrate how planning, candidate selection, and revision behaviors vary independently across models.

\paragraph{Workflow structure and production effort.}
A median workflow graph contains 121 nodes and 131 edges, with an artifact dependency depth of nine, reflecting substantial multi-stage coordination across generation, inspection, and editing.
As shown in \cref{tab:process}, workflow sizes vary considerably across models, ranging from 54 to 160 nodes and depths of 7 to 11.
The share of production nodes reaching the final film ranges from 72\% to 95\%, indicating different degrees of candidate exploration and selection throughout production.
For instance, Opus 5.5 and DeepSeek V4.1 Flash construct larger workflows (160 and 150 nodes) but retain only 75\% and 72\% of their production nodes, whereas MiMo and Doubao retain 95\% and 93\%, respectively.
In terms of production tools, video generation supplies 74--92\% of delivered shots, with the remainder coming from code rendering and image animation; product tasks rely most heavily on code rendering (41\%).
Execution effort also differs substantially across models: median task duration ranges from 47 to 125 minutes, while input tokens range from 0.51M for MiMo to 6.39M for Opus 5.5.

\begin{table}[ht]
\centering
\caption{\textbf{Workflow graph and process statistics, used for attribution only.} Medians per task: tool calls, depth of the film in the graph, share of production nodes that reach the film, minutes, and tokens (input with cache / output).}
\vspace{2mm}
\label{tab:process}
\footnotesize
\setlength{\tabcolsep}{1.5pt}
\begin{tabular*}{\linewidth}{@{\extracolsep{\fill}}lrrrrr@{}}
\toprule
Model & Nodes & Depth & Used & Min. & Tokens \\
\midrule
Claude Opus 5.5 & 160 & 11 & 0.75 & 125 & 6.39M / 149k \\
Gemini 4 Argon & 140 & 9 & 0.84 & 79 & 1.73M / 47k  \\
DeepSeek V4.1 Flash & 150 & 10 & 0.72 & 81 & 4.21M / 82k  \\
Qwen3.8 Max & 97 & 9 & 0.85 & 79 & 1.65M / 47k \\
GPT-6 Astra & 127 & 10 & 0.88 & 56 & 0.57M / 16k \\
MiMo v2.6 Pro & 76 & 7 & 0.95 & 66 & 0.51M / 54k  \\
Doubao Seed 2.1 Pro & 54 & 7 & 0.93 & 47 & 0.58M / 18k \\
\bottomrule
\end{tabular*}
\vspace{-3mm}
\end{table}

Together, these results demonstrate that general MLLMs can autonomously establish a common multi-stage production structure, yet differ substantially in workflow complexity, candidate selection, and resource allocation. Such differences highlight the flexibility of agentic audiovisual production and motivate further investigation into how production decisions relate to final film quality.


\subsection{RQ3: Capability Bottlenecks}
\label{sec:rq3}
\vspace{2mm}
\begin{boxA}
\noindent\textbf{Takeaway:}
Agentic MLLMs can execute workflows and recover from explicit errors, but struggle to detect and correct perceptual defects. Higher quality is associated with more selective production rather than additional tool calls, highlighting the missing quality loop of effective inspection, selection, and revision.
\end{boxA}
\vspace{2mm}
\noindent\textbf{From workflow execution to capability diagnosis.}
RQ1/2 show that MLLMs organize production workflows and deliver audiovisual works, yet quality gaps remain.
We pair high- and low-scoring runs on the same task and analyze their workflow graphs and evaluation evidence, identifying failure points and underlying capabilities. Independently, \cref{tab:errors} quantifies four error modes using rule-based proxies across the delivered films. Models recover from 80--90\% of reported tool errors, suggesting that the remaining bottlenecks extend beyond conventional error handling.

\begin{table}[t]
\centering
\caption{\textbf{Capability-related error modes.} Prevalence (\%) of four rule-based error proxies across the 1{,}063 delivered films, with the lowest and highest rates among models.}
\label{tab:errors}
\footnotesize
\setlength{\tabcolsep}{0pt}
\begin{tabular*}{\linewidth}{@{\extracolsep{\fill}}llrll@{}}
\toprule
Error mode & Ability & All & Lowest & Highest \\
\midrule
Character drift & keeping state & 19.1 & 10.9 Claude & 27.5 Doubao \\
AV misalignment & keeping state & 32.9 & 28.2 Claude & 37.5 Qwen \\
Silent damage & verifying outputs & 3.5 & 0.6 Gemini & 5.8 MiMo \\
Tool-length pacing & deciding by effect & 42.1 & 25.6 Claude & 59.4 Doubao \\
\bottomrule
\end{tabular*}
\end{table}

\noindent\textbf{Gap 1: Maintaining state across stages.}
Character consistency and audiovisual synchronization require information to persist across multiple generation and editing steps.
As shown in \cref{tab:errors}, character drift and audiovisual misalignment remain recurring production defects.
Maintaining a consistent character, for example, requires reference images, image edits, video generation and subsequent audio synchronization to preserve compatible states.
These failures highlight the difficulty of coordinating interdependent outputs throughout multi-stage production.

\begin{figure}[t]
\centering
\includegraphics[width=\linewidth]{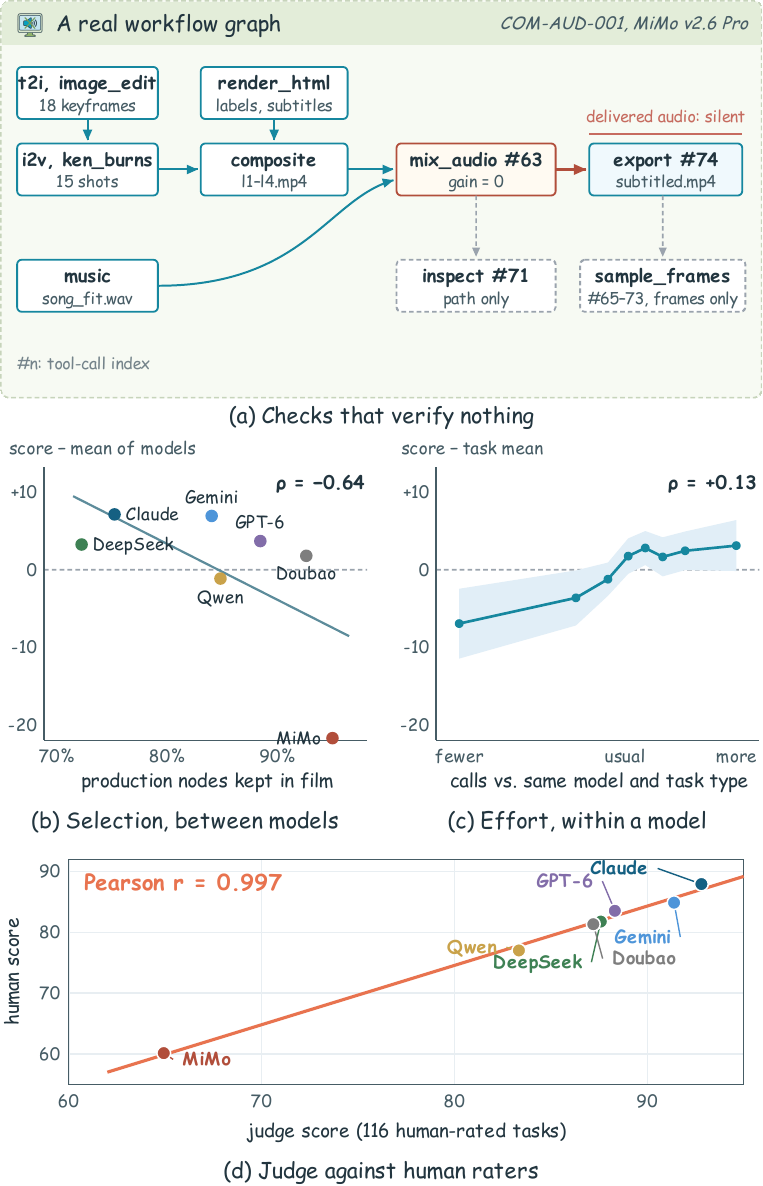}
\caption{\textbf{The missing quality loop in audiovisual production.}
(a) A traced run of a silent film: the final mixing call sets the gain to zero, and the later checks only sample frames or list the file.
(b) Share of production nodes that reach the film against the paired score, between models. (c) Paired score against relative tool calls within a model.
(d) Agreement between automated judge scores and human ratings across 116 tasks.
}
\vspace{-2mm}
\label{fig:effort}
\end{figure}

\noindent\textbf{Gap 2: Verifying outputs beyond tool success.}
Models often fail to recognize defects that produce no explicit tool error.
Delivered films are occasionally silent (\cref{tab:defects}); in four of the silent runs we traced (MiMo v2.6 Pro's COM-AUD-001, CRE-AUD-001, MUS-OPN-003 and MUS-REF-001), the final mixing call sets the gain to zero (\cref{fig:effort}(a)), and the checks that follow sample frames or list the file rather than listen to the result, so no corrective production step follows.
This exposes a gap between confirming artifact existence and verifying its actual perceptual quality.

\noindent\textbf{Gap 3: Selecting and revising by observed effect.}
Additional production effort alone does not guarantee higher quality.
Within each model, tool-call counts show only a weak association with film quality ($\rho=+0.13$ across 1{,}120 runs; \cref{fig:effort}(c)).
In contrast, the fraction of production nodes retained in the final film is negatively correlated with quality across models ($\rho=-0.64$; \cref{fig:effort}(b)), while workflow depth shows a positive correlation ($\rho=+0.75$).
These findings suggest that higher-quality models tend to explore deeper workflows, generate alternatives and selectively retain useful outputs, rather than simply increasing tool usage.





\noindent\textbf{The quality loop remains the central bottleneck.}
Reliable production requires more than understanding instructions, organizing workflows and invoking tools; models must also maintain state, detect perceptual defects and revise based on their effects. Closing this quality loop is essential for reliable autonomous audiovisual production.

\noindent\textbf{Human validation of the automated judge.}
Across 116 human-rated tasks, judge scores correlate with the mean of three human raters at Pearson $r=0.997$ across the seven models and order all 21 model pairs identically (\cref{fig:effort}(d)). Detailed human evaluation results are provided in Appendix~\ref{sec:supp_human}.

\section{Discussion and Conclusion}
\vspace{-2mm}
\label{sec:conclusion}

\sbench makes a foundation model the director of productive films and gives it nothing but atomic tools.
At the level of the product, explicit requirements are close to saturated, while quality still differs clearly.
At the level of the process, every model we tested builds a complete workflow graph by itself.
At the level of capability, models finish films reliably but make them good unreliably, since the quality loop of state, verification and effect-driven decisions is missing.
These findings suggest that progress depends not simply on expanding tool use, but on enabling agents to use perceptual feedback to guide subsequent production decisions.
By jointly evaluating finished films and their underlying workflows, \sbench provides a foundation for studying and advancing quality-aware autonomous audiovisual creation.



\clearpage
\section*{LLM Usage Statement}

OpenAI Codex was used to assist with manuscript preparation,
including drafting and revising the conclusion, adapting the
existing \LaTeX{} project to the CVPR template, formatting
tables and figures, and resolving \LaTeX{} compatibility issues.

{
    \small
    \bibliographystyle{ieeenat_fullname}
    \bibliography{refs_found,refs_models}
}

\clearpage
\appendix
\setcounter{page}{1}
\maketitlesupplementary

\section{Task Sources and Genre Matrix}
\label{sec:supp_tasks}
\noindent\textbf{Sources.}
Every task is adapted from an AIGC video published on a public video platform.
Of the 160 source videos, 129 come from Bilibili, 19 from LibTV, 10 from AcFun and 2 from PixTV; the screening model judged 155 of them to be AI-generated and did not judge the other 5.
The source videos have a median length of 118\,s.
Annotators assigned each candidate a genre and a task type, and gave it a one-sentence brief.
For most candidates an LLM first wrote three candidate briefs from the video, and the annotator chose one, edited it or wrote a new one.
Of the 160 briefs, 127 were taken from such candidates (written by GPT-6.1 Sol, Claude Opus 5.5, Gemini 3.1 Pro and Kimi K3.1 Pro), 30 come from an earlier version of the same pipeline and 3 were written by annotators.

\noindent\textbf{Screening.}
Candidates were checked automatically against a written screening standard: the source video is available, at least 30\,s long, has an audio track when sound is required, has a short side of at least 720 pixels and is not a duplicate; the brief states a duration and an aspect ratio, is not a shot-by-shot prompt, names no other generation model or production pipeline, and does not name a real brand, IP or person.
The checks were run after annotation and recorded per task; they did not remove tasks automatically.
Among the 160 tasks the most frequent exceptions are a brief longer than three sentences (40), a real name in the task description (29) and a source video whose short side is below 720 pixels (12).

\noindent\textbf{Sampling.}
From 1{,}275 usable candidates we drew 20 tasks per genre.
Within a genre the 20 places are spread evenly across the five task types, and a type with fewer candidates than its share gives the remainder to the others.
Within each genre and task type we kept the candidates with the highest screening score, the mean of the available screening votes and the review score of the brief.
\Cref{tab:genre_matrix} lists the sampled tasks and the candidate pool for every genre and task type.

\noindent\textbf{Conditioned task types.}
In this release, Asset-conditioned and Reference-conditioned tasks state their condition in the brief only.
No asset or reference file is provided, so all 160 user messages have the same three fields (\cref{sec:supp_tools}).

\begin{table}[h]
\centering
\caption{\textbf{Genre by task type.} Sampled tasks / candidates in the pool.}
\label{tab:genre_matrix}
\footnotesize
\setlength{\tabcolsep}{2.5pt}
\begin{tabular*}{\linewidth}{@{\extracolsep{\fill}}lrrrrrr@{}}
\toprule
Genre & Open & Constr. & Asset & Ref. & Audio & All \\
\midrule
Narrative & 4 / 65 & 4 / 119 & 4 / 5 & 4 / 7 & 4 / 5 & 20 / 201 \\
Commercial & 4 / 54 & 4 / 116 & 4 / 40 & 4 / 6 & 4 / 5 & 20 / 221 \\
Product & 0 / 0 & 12 / 29 & 8 / 8 & 0 / 0 & 0 / 0 & 20 / 37 \\
Education & 5 / 59 & 5 / 148 & 5 / 18 & 1 / 1 & 4 / 7 & 20 / 233 \\
Music & 4 / 6 & 4 / 4 & 4 / 14 & 4 / 4 & 4 / 251 & 20 / 279 \\
Documentary & 8 / 23 & 8 / 69 & 2 / 2 & 0 / 0 & 2 / 2 & 20 / 96 \\
Creator & 6 / 16 & 7 / 30 & 4 / 4 & 2 / 2 & 1 / 1 & 20 / 53 \\
Dialogue & 1 / 1 & 15 / 150 & 3 / 3 & 0 / 0 & 1 / 1 & 20 / 155 \\
\midrule
All & 32 / 224 & 59 / 665 & 34 / 94 & 15 / 20 & 20 / 272 & 160 / 1275 \\
\bottomrule
\end{tabular*}
\end{table}

\begin{figure*}[h]
\centering
\includegraphics[width=\linewidth]{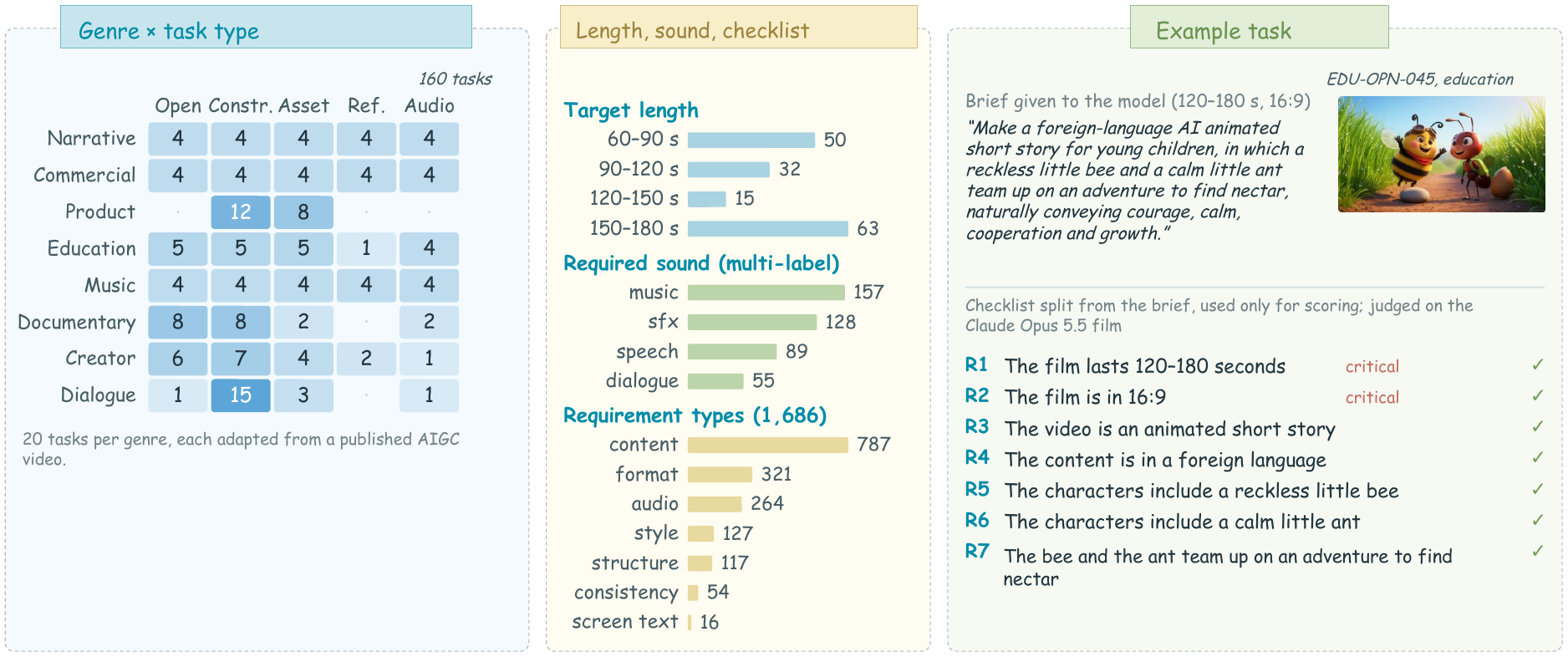}
\caption{\textbf{Published films, eight genres, judgeable requirements.}
Left: genre by task type for the 160 tasks (Constr.\ = Constraint, Ref.\ = Reference-conditioned, Audio = Audio-led). Middle: target lengths, required sound types (multi-label) and content types of the 1{,}686 checklist requirements. Right: the brief of task EDU-OPN-045 and its full checklist, which is used only for scoring; every item is met by the film of Claude Opus 5.5.}
\label{fig:dataset}
\end{figure*}

\section{Requirement Checklists}
\label{sec:supp_checklist}
\noindent\textbf{Splitting.}
An LLM splits each user message into a checklist and sees only that message, never the source video or any reference.
It lists only requirements that the message states explicitly and that can be verified on the final film, each as the smallest unit that can be judged on its own.
These cover content, events, characters and their relations, captions, lines and sound, product and delivery requirements, and prohibitions.
It may not turn aesthetic preferences into requirements, add common-sense or reference details, or list what the brief leaves free.
Every item must quote a contiguous span of the message as its source, and the span is checked by exact string match.
The duration window and the aspect ratio are added by the program as the first two items.

\noindent\textbf{Critical items.}
An item is critical only when the brief states it as a must, a prohibition or a required ending; the duration and the aspect ratio are always critical.
A critical item that fails flags the film for review but carries no extra penalty.
The 160 checklists have 1{,}686 items, 4--18 per task with a median of 10, and 545 of them are critical.

\noindent\textbf{Judging.}
The judge rates every item as met (1), partly met (0.5), not met (0), not applicable or unobservable, with evidence and timestamps, and judges the duration and the aspect ratio from the measured metadata of the file.
Requirement completion is the mean over applicable items.
\Cref{fig:dataset} shows the checklist of EDU-OPN-045, and \cref{tab:checklists} gives three more, translated from Chinese; critical items are marked with $\ast$.

\begin{table*}[t]
\centering
\caption{\textbf{Full checklists of three tasks} (translated). $\ast$: critical item.}
\label{tab:checklists}
\footnotesize
\begin{tabularx}{\linewidth}{@{}>{\raggedright\arraybackslash}X>{\raggedright\arraybackslash}X>{\raggedright\arraybackslash}X@{}}
\toprule
\textbf{PRO-CON-026} (Product, Constraint) & \textbf{DOC-OPN-019} (Documentary, Open Brief) & \textbf{COM-AST-002} (Commercial, Asset-conditioned) \\
\emph{Make a one-minute 16:9 landscape 3D advertisement for the fictional brand Aervane: in a futuristic lab, an electric shaver assembles itself from loose parts, shows its cutting head and magnetic structure, and ends with a head-shaving demonstration and a high-pressure water-jet waterproof test. Black-grey metal with electric blue, cool and premium, with electronic music and mechanical sound effects, no human voice.}
&
\emph{Make a one-minute 16:9 landscape realistic nature documentary on the rhythm of life on Earth. Open with misty forests, crashing waves and polar scenery to show the flow of time, turn in the middle to predation and escape on grasslands and lakes, and end with macro close-ups of insects and plants and aerial shots of the wilderness, closing on the image of a ``web of life''. Use a deep male Mandarin narration at a calm pace over grand orchestral music and natural sounds such as wind and waves. The picture should feel like real wildlife and aerial photography; species and locations are free.}
&
\emph{Make a 1--2 minute 16:9 landscape concept advertisement for a real-estate project in Dongguan, blending realistic architecture, aerial views and digital painting, bright and uplifting. Open with city traffic and high-rises, link landmark culture with a golden basketball and floating red lanterns, move into family, sports and home life, and close with the seasons and the project reveal. The Dongguan Basketball Center and Qifeng Park must appear and the ending must clearly show the project name; other landmarks and transitions are free, and the building must look the same throughout. No narration, dialogue or singing; driving electronic music with strings and a few whoosh transitions.} \\
\midrule
R1$^\ast$ Duration 60--65\,s. \newline
R2$^\ast$ Aspect ratio 16:9. \newline
R3 Includes elements of the fictional brand Aervane. \newline
R4 A 3D advertisement. \newline
R5 Set in a futuristic lab. \newline
R6 The shaver assembles itself from loose parts. \newline
R7 Shows the cutting head and the magnetic structure. \newline
R8$^\ast$ Ends with a head-shaving demonstration and a high-pressure water-jet waterproof test. \newline
R9 Black-grey metal with electric blue. \newline
R10 Electronic music and mechanical sound effects. \newline
R11$^\ast$ No human voice in the audio.
&
R1$^\ast$ Duration 63--73\,s. \newline
R2$^\ast$ Aspect ratio 16:9. \newline
R3 A realistic nature documentary. \newline
R4 The theme is the rhythm of life on Earth. \newline
R5 Opens with misty forests, crashing waves and polar scenery. \newline
R6 The middle shows predation and escape on grasslands and lakes. \newline
R7$^\ast$ The ending shows macro close-ups of insects and plants. \newline
R8$^\ast$ The ending includes aerial shots of the wilderness. \newline
R9 A deep male Mandarin narration at a calm pace. \newline
R10 A grand orchestral score. \newline
R11 Natural sounds such as wind and waves.
&
R1$^\ast$ Duration 77--87\,s. \newline
R2$^\ast$ Aspect ratio 16:9. \newline
R3 Opens with city traffic and high-rises. \newline
R4 A golden basketball appears. \newline
R5 Floating red lanterns appear. \newline
R6 Shows family, sports and home life. \newline
R7 Shows the change of seasons. \newline
R8$^\ast$ The Dongguan Basketball Center appears. \newline
R9$^\ast$ Qifeng Park appears. \newline
R10$^\ast$ The ending clearly shows the project name as a caption. \newline
R11 The project building looks the same throughout. \newline
R12$^\ast$ No narration, dialogue or singing. \newline
R13 Electronic music with strings. \newline
R14 A few whoosh transition effects. \\
\bottomrule
\end{tabularx}
\end{table*}

\section{Tools and System Prompt}
\label{sec:supp_tools}
\paragraph{Agent loop.}
Every model runs in the same agent loop, Claude Code in non-interactive mode, connected to the model through an Anthropic-compatible gateway at its highest reasoning-effort setting.
The default system prompt of Claude Code is replaced by the prompt below, its built-in tools (shell, file reading and writing, web access) are disabled, and user settings, memory files and plug-ins are not loaded.
The model therefore sees only the system prompt, one user message and the 33 tools of \cref{tab:tools}, served by one MCP server.
A further entry, \texttt{batch}, submits up to 16 independent calls to run in parallel; each of them counts against the budget like a separate call.
Tools that return pictures (\texttt{sample\_frames}, \texttt{scene\_detect}, the previews of the rendering tools, \texttt{remove\_bg}, \texttt{color} on an image and images from \texttt{fetch\_url}) attach them to the result as images, and every result reports the calls used and the minutes left.

\paragraph{User message.}
The user message has three fields: the brief, the duration window and the aspect ratio.
The environment can also pass required assets, references and given audio, copied into \texttt{inputs/}, but no task in this release uses them (\cref{sec:supp_tasks}).
For task EDU-OPN-045 (\cref{fig:dataset}) it reads, in translation: \emph{Request: make a foreign-language AI animated short story for young children, in which a reckless little bee and a calm little ant team up on an adventure to find nectar, naturally conveying courage, calm, cooperation and growth. Duration: 120--180 seconds. Aspect ratio: 16:9.}

\begin{table*}[t]
\centering
\caption{\textbf{The 33 atomic tools.} Arguments in brackets are optional; every tool that writes a file also takes an optional output path. Backends are fixed for all runs.}
\label{tab:tools}
\scriptsize
\setlength{\tabcolsep}{4pt}
\begin{tabularx}{\linewidth}{@{}l>{\raggedright\arraybackslash}p{4.9cm}>{\raggedright\arraybackslash}X>{\raggedright\arraybackslash}p{2.4cm}@{}}
\toprule
Tool & Arguments & Function & Backend \\
\midrule
\multicolumn{4}{@{}l}{\emph{Video generation}} \\
\texttt{t2v} & prompt [duration, aspect, seed] & Text to video; 5\,s or 8\,s, short side 768, with its own audio track & MiniMax-H3 \cite{minimax2026h3} \\
\texttt{i2v} & prompt, first\_frame [last\_frame, duration, seed] & Video from a first frame and an optional last frame; aspect follows the frame & MiniMax-H3 \\
\texttt{ref\_video} & prompt, references [duration, aspect, seed] & Video conditioned on up to 9 images and 3 videos, cited in the prompt as \textless Image 1\textgreater, \textless Video 1\textgreater & MiniMax-H3 \\
\multicolumn{4}{@{}l}{\emph{Image}} \\
\texttt{t2i} & prompt [aspect, seed, steps] & Character sheets, scenes and keyframes, sized to serve as first frames & Qwen-Image \cite{wu2025qwenimage} \\
\texttt{image\_edit} & src, prompt [ref, seed] & Edit an image by text with an optional reference image, e.g.\ to keep a character & Qwen-Image-Edit-2509 \\
\texttt{remove\_bg} & image [model] & Cut out the subject as a transparent PNG & IS-Net \cite{qin2022dis} \\
\multicolumn{4}{@{}l}{\emph{Sound}} \\
\texttt{tts} & text [voice\_ref, voice\_ref\_text, speaker, instruct, speed, language] & Narration and dialogue with nine preset voices, style instructions or voice cloning from a reference clip & Qwen3-TTS \cite{hu2026qwen3tts} \\
\texttt{music} & prompt [duration, lyrics] & Music of 3--300\,s from a description and optional lyrics & ACE-Step 1.5 \cite{gong2026acestep15} \\
\texttt{sfx} & video, prompt [duration, negative\_prompt] & Action and ambient sound generated for a given shot & MMAudio \cite{cheng2025mmaudio} \\
\texttt{lipsync} & video, audio & Re-render a speaking face to match a speech track & LatentSync 1.6 \cite{li2024latentsync} \\
\multicolumn{4}{@{}l}{\emph{Code rendering} (silent output; \texttt{times} or \texttt{frames} return preview stills instead of a video)} \\
\texttt{render\_html} & entry or html [duration, fps, width, height, transparent, times] & Render an HTML, CSS, SVG, Canvas or WebGL page frame by frame through \texttt{window.render(t)} or a virtual clock; offline libraries such as three.js, GSAP, D3 and KaTeX & Headless Chromium \\
\texttt{render\_three} & code or entry [duration, fps, width, height, transparent, times] & A three.js r170 scene module that returns \texttt{update(t)} & Headless Chromium \\
\texttt{render\_blender} & code or script [frames, start, end, engine, samples, transparent] & A bpy scene script rendered with Cycles, EEVEE or Workbench, at most 60\,min per call & Blender 5.2.0 \\
\multicolumn{4}{@{}l}{\emph{Editing}} \\
\texttt{trim} & video, start, end & Cut a segment, frame-accurate & FFmpeg \\
\texttt{concat} & videos & Join clips in order, unifying resolution and frame rate & FFmpeg \\
\texttt{transition} & videos [kind, duration] & Join clips with one of over 30 transitions and an audio crossfade & FFmpeg \\
\texttt{overlay} & video [subtitles, srt, srt\_text, font\_size] & Burn in subtitles & FFmpeg \\
\texttt{composite} & video, layer [start, end, position, x, y, width, opacity, chroma\_key] & Overlay an image or video layer with alpha or a chroma key: titles, logos, end cards, picture-in-picture & FFmpeg \\
\texttt{reframe} & video, width, height [mode, anchor, fps] & Change aspect ratio, resolution or frame rate by crop, pad, blurred fill or stretch & FFmpeg \\
\texttt{speed} & video [factor, reverse] & Change speed with pitch preserved, or reverse & FFmpeg \\
\texttt{color} & video [preset, brightness, contrast, saturation, gamma, temperature, lut, vignette] & Grade a clip or image with eight presets, adjustments or a LUT & FFmpeg \\
\texttt{ken\_burns} & image, duration [zoom\_from, zoom\_to, pan] & Pan and zoom a still image into a silent clip & FFmpeg \\
\multicolumn{4}{@{}l}{\emph{Audio processing}} \\
\texttt{mix\_audio} & video, audio [gain, replace, start] & Mix a track into the soundtrack of a video or replace it; \texttt{gain} is a linear factor and the output keeps the duration of the video & FFmpeg \\
\texttt{audio\_mix} & tracks [duck, duck\_db, normalize\_lufs, duration] & Mix tracks with per-track start, gain in dB, fades and trims, ducking under speech and loudness normalization & FFmpeg \\
\texttt{audio\_edit} & audio [start, end, fade\_in, fade\_out, gain\_db, normalize\_lufs, speed, pad\_to, denoise] & Edit one track & FFmpeg \\
\multicolumn{4}{@{}l}{\emph{Inspection and search}} \\
\texttt{sample\_frames} & video [fps, start, end] & Contact sheet of time-stamped frames, returned as an image; an image file is returned directly & FFmpeg \\
\texttt{inspect} & path & Duration, resolution, frame rate and audio of a file, or the listing of a directory & FFmpeg \\
\texttt{asr} & audio [language] & Time-stamped transcript of speech & Whisper large-v3 \cite{radford2023whisper}, Chinese fine-tune \\
\texttt{scene\_detect} & video [threshold] & Shot boundaries and a contact sheet of first frames & FFmpeg \\
\texttt{beat\_detect} & audio [start, end] & Tempo, beats and strongest accents & librosa \\
\texttt{search} & query [max\_results] & Up to five web results with title, URL and snippet; pages of the source video are removed & Web search \\
\texttt{fetch\_url} & url [max\_mb] & Download an image, audio, video, page or 3D model (200\,MB by default); images are shown and pages return their text; the source video is blocked & HTTP \\
\multicolumn{4}{@{}l}{\emph{Delivery}} \\
\texttt{export} & src & Copy a video to \texttt{final.mp4} as the delivered film; it may be called again, and the last call counts & FFmpeg \\
\bottomrule
\end{tabularx}
\end{table*}

\paragraph{System prompt.}
The system prompt is the same for every model and task; only the budget at its end is filled in, with 200 calls and 4 or 8 hours (\cref{sec:task}).
It describes each tool and attaches usage advice to several of them: a character sheet before editing to keep a person, preview stills before a full render, a look at the frames after every generation and a check of duration and audio before export. It gives no plan for a whole film, such as which steps to take or in what order to script, generate, edit and mix.
The original is in Chinese; we give a faithful translation, with tool names, arguments and file names unchanged.

\newcommand{\pl}[1]{\par\hangindent=1em\hangafter=1 -- #1}
\begin{tcolorbox}[sbprompt,title={System prompt (translated from Chinese)}]
You are a video director. Use the tools below to make beautiful, creative videos. Write the script and the storyboard in your replies.

Stories and explainers need at least 3 shots. Videos generated by H3 (\texttt{t2v} / \texttt{i2v} / \texttt{ref\_video}) may only be 5 or 8 seconds per segment; clips from code rendering or image pan-and-zoom have no length limit. Join longer films with \texttt{concat} or \texttt{transition}.

Picture sources:
\pl{\texttt{t2v}: text to video, H3 fl2va}
\pl{\texttt{i2v}: video from a first frame, fl2va. Pass a last frame too if you have one}
\pl{\texttt{ref\_video}: video from reference images or reference videos, ref2va}
\pl{\texttt{t2i}: character sheets, scenes, keyframes}
\pl{\texttt{image\_edit}: edit images. To keep the same person, make a character sheet first, edit it, then \texttt{i2v} or \texttt{ref\_video}}
\pl{\texttt{render\_html}: renders an HTML / CSS / SVG / Canvas / WebGL page frame by frame into a video (or a video with a transparent background). If the page defines \texttt{window.render(t)}, it is drawn frame by frame through it; otherwise ordinary animation (CSS animation, requestAnimationFrame, GSAP) is recorded with a virtual clock. Suited to titles, lower thirds, data graphics, motion graphics and explainer animation}
\pl{\texttt{render\_three}: write a 3D scene in three.js that returns \texttt{update(t)}, rendered into a video}
\pl{\texttt{render\_blender}: build a scene with a Blender 5.2.0 Python (bpy) script and render it into a video or stills (Cycles / EEVEE / Workbench)}
\pl{\texttt{ken\_burns}: pan and zoom a still image into a video; \texttt{fetch\_url} can download real images and material; \texttt{remove\_bg} cuts out images}
\pl{Videos from code rendering are all silent; make the sound with the tools below, then \texttt{mix\_audio}. Render slow long shots in segments, then \texttt{concat} or \texttt{transition}. First render a few stills with \texttt{times} (\texttt{render\_html} / \texttt{render\_three}) or \texttt{frames} (\texttt{render\_blender}) to confirm the composition, then render the whole segment}

Sound:
\pl{\texttt{tts}: narration and dialogue. With a reference voice, pass the reference audio as the prompt wav (\texttt{voice\_ref})}
\pl{\texttt{music}: score}
\pl{\texttt{sfx}: action and ambient sound, generated from the video of the shot}
\pl{\texttt{lipsync}: when someone on screen speaks this line, lip-sync the video to the \texttt{tts} audio}
\pl{\texttt{asr}: check your own narration and dialogue}
\pl{\texttt{audio\_mix}: multitrack mixing (start, gain, fades, ducking under speech, loudness normalization); \texttt{audio\_edit}: trimming, fades, gain and speed of a single track}
\pl{\texttt{beat\_detect}: BPM and beats of music, for cutting on the beat}

Editing:
\pl{\texttt{trim}, \texttt{concat}, \texttt{transition} (joining with transitions), \texttt{overlay} (burned-in subtitles), \texttt{composite} (layers: titles, logos, end cards, picture-in-picture, with alpha)}
\pl{\texttt{reframe} (aspect ratio and resolution), \texttt{speed} (speed change, reverse), \texttt{color} (grading)}
\pl{\texttt{mix\_audio}: mix an audio track into the soundtrack of a video or replace it. When voice and music play together, the voice must be intelligible}

Inspection and material:
\pl{\texttt{sample\_frames}: look at the picture after every generation or render. If it is not good, change it and generate again; do not call again with exactly the same arguments}
\pl{\texttt{scene\_detect}: shot segmentation of a reference film}
\pl{\texttt{search}: for education, documentary and product facts, search before writing the narration. It returns material, not scoring criteria}
\pl{\texttt{inspect}: before submitting, check duration, aspect ratio and audio track}
\pl{\texttt{export}: the only way to submit. Without \texttt{export}, nothing is submitted}

Parallelism: generation tools (images, image editing, video, voice, effects, music, rendering) are slow, so do not queue independent jobs one after another. Submit them together with \texttt{batch} (for example all keyframes in one batch, then all image-to-video jobs in another), or issue several tool calls in the same reply; they run in parallel and return together when all are done. When a step needs the output of an earlier one, wait for that one to return before issuing it. Every call counts against the budget as usual.

Working directory: the current directory is the working directory; the assets, references and audio given by the user are under \texttt{inputs/}. Put all outputs in the working directory and pass relative paths to the tools. Work only in the working directory and do not read or write evaluation data outside it (other runs, the task sheet and so on), or this run is void.

Do not ask the user for scoring criteria. For assets and references use only the files the user gave, or things you made or found yourself.

Budget: \{max\_calls\} tool calls, \{wall\_hours\} hours. When time is up, stop with the film you have already exported. The \texttt{budget} field in every tool result gives the calls used and the minutes left.
\end{tcolorbox}

\section{Checklist Judging and Scoring}
\label{sec:supp_metrics}
\noindent\textbf{Calls.}
The judge is Gemini 3.8 Flash at temperature 0.
For each film and pass it is called six times in a fixed order.
(1) \emph{Listening}: from the audio track alone, it lists every speech segment with speaker, language, foreground or background role and the words it can make out, together with music, effects, silences of more than one second and odd sounds.
(2) \emph{Sound}: from the audio track, it answers the 21 sound checks.
(3) \emph{Expression}: from the picture and the audio, it first summarizes in one or two sentences what the film says and then answers the 7 expression checks, the last of which is specific to the genre.
(4) \emph{Summary check}: from text alone, it judges whether that summary captures the main point of the brief.
(5) \emph{Picture}: from the picture, 12 high-resolution frames of the original file and face crops, it answers the 29 picture checks.
(6) \emph{Audio-visual}: from the picture, the audio, mouth crops of every speech segment and a table of measured mouth movement, it answers the 16 audio-visual checks.
A silent film skips the listening, sound and audio-visual calls.
Requirement completion is judged item by item by the same judge in a separate call that sees the checklist of \cref{sec:supp_checklist}.

\noindent\textbf{Measurements.}
Before judging, a program measures integrated loudness, short-term loudness range, channel balance, clipping, silence and silence in the last 40\% of the film; blur, black, frozen and shaking frames; and, in every speech segment, the share of frames with a detected face (YuNet), the variation of mouth opening from 68 facial landmarks (LBF) and its correlation with speech energy.
Suspicious segments become numbered candidates in the matching call, and the judge must confirm each one, with the check it belongs to, or reject it with a reason such as a normal cut, an effect asked for by the brief or a face seen in profile.

\noindent\textbf{Validation.}
Every answer is checked by program: time spans must be whole milliseconds within the film, every check must be answered, every candidate resolved, and a not-applicable answer must quote the brief or state that the object is absent.
An invalid answer is returned to the judge with the problems listed and asked again once.

\noindent\textbf{Costs.}
\Cref{tab:metrics} lists the checks of each metric and their fatal conditions.
A present check that is not fatal costs 1 for a single instance, 1.5 for several and 2 when its instances cover more than 30\% of the film.
Seven checks are minor and cost 0.25 each, at most 1 per metric: a loop seam in the music, noise or flicker, slight deformation of a background figure or object, ambience that does not fit the scene, music cut abruptly at an edit, slight differences in a character's face between shots, and subtitles that do not cover the subject.
The program adds its own costs: integrated loudness outside $-30$ to $-9$\,LUFS or a short-term loudness range above 15\,LU costs 1 on mix balance, a channel imbalance above 6\,dB costs 0.25, and a speech rate above 6.5 or below 2 Chinese characters per second costs 1 on intelligibility.
When the brief asks for speech and the film is silent for more than 30\% of its length, or for more than 80\% of its last 40\%, every sound and audio-visual metric takes the lowest level.
A summary that misses the main point of the brief costs 1 on expression and makes ``no clear message'' fatal.

\noindent\textbf{Levels and calibration.}
A metric takes the lowest level when a fatal check appears in at least half of the passes, here in at least one of the two; otherwise its cost, averaged over the passes, is compared with the thresholds $a$ and $b$ of \cref{tab:cuts}.
The thresholds were chosen on the 27 development tasks of \cref{sec:supp_human} from a grid of 0 to 4, maximizing the mean quadratic-weighted $\kappa$ with each rater minus 0.1 times the $L_1$ distance between the machine and human level distributions.
The judge sees neither the costs nor the thresholds.

\noindent\textbf{Aggregation.}
The three levels count as 0, 0.5 and 1.
Within semantics, requirement completion and expression are weighted 2:1; sound, picture and audio-visual alignment average their applicable metrics.
The overall score weights the four dimensions by the genre of the task (\cref{tab:cuts}), renormalizing when a dimension has no applicable metric.

\begin{table}[h]
\centering
\caption{\textbf{Thresholds and genre weights.} Left: a cost of at most $a$ gives the top level and at most $b$ the middle level. Right: weights (\%) of sound, semantics, picture and audio-visual alignment.}
\label{tab:cuts}
\footnotesize
\setlength{\tabcolsep}{3pt}
\begin{tabular}{@{}lrr@{\hspace{10pt}}lrrrr@{}}
\toprule
Metric & $a$ & $b$ & Genre & AUD & SEM & VIS & AV \\
\midrule
AUD-C1 & 1.5 & 2.5 & Narrative & 20 & 30 & 25 & 25 \\
AUD-C2 & 0.5 & 2 & Commercial & 20 & 30 & 30 & 20 \\
AUD-C3 & 0.5 & 1 & Product & 15 & 35 & 35 & 15 \\
AUD-C4 & 1 & 1.5 & Education & 25 & 40 & 20 & 15 \\
SEM-C2 & 0.5 & 1.5 & Music & 40 & 10 & 20 & 30 \\
VIS-C1 & 0.25 & 2 & Documentary & 25 & 35 & 25 & 15 \\
VIS-C2 & 0 & 1 & Creator & 25 & 30 & 25 & 20 \\
VIS-C3 & 0 & 1 & Dialogue & 30 & 35 & 15 & 20 \\
VIS-C4 & 0.5 & 1.5 & & & & & \\
AV-C1 & 0.5 & 2 & & & & & \\
AV-C2 & 3 & 3 & & & & & \\
AV-C3 & 1.5 & 4 & & & & & \\
\bottomrule
\end{tabular}
\end{table}

\begin{table*}[t]
\centering
\caption{\textbf{Thirteen whole-film metrics and their defect checks} (translated). Requirement completion is the mean over checklist items; every other metric takes one of three levels from its checks. Fatal conditions set the lowest level; \emph{on the subject} means on the main person or object of the shot.}
\label{tab:metrics}
\footnotesize
\begin{tabularx}{\linewidth}{@{}l>{\raggedright\arraybackslash}p{2.6cm}r>{\raggedright\arraybackslash}X@{}}
\toprule
ID & Metric & Checks & What is checked (\emph{fatal conditions in italics}) \\
\midrule
SEM-C1 & Requirement completion ($R$) & -- & Every checklist item: met 1, partly met 0.5, not met 0; not-applicable items excluded; mean of the applicable items. \\
SEM-C2 & Expression (named per genre) & 7 & No clear main message (\emph{fatal if the summary also misses the brief}); jumps between sections; dragging or rushed pacing; contradictions or checkable factual errors; no ending; subtitle errors; a genre check, \eg broken cause and effect for Narrative or unclear selling points for Commercial. \\
\midrule
VIS-C1 & Clarity and composition & 7 & Blurred subject (\emph{fatal on the subject for over 30\% of the film}); corrupted texture (\emph{fatal on the subject}); ghosting (\emph{fatal on the subject}); unexplained marks on the subject (\emph{fatal on the subject}); over- or under-exposure; cropped or blocked subject; noise or flicker. \\
VIS-C2 & Body and object structure & 8 & Extra or missing limbs, distorted or melted faces (\emph{both fatal on the subject}); a malformed mouth while speaking (\emph{fatal twice on the subject}); malformed hands; physically impossible spatial relations (\emph{fatal on the subject}); wrong object structure; sudden scale changes or impossible twists; slight deformation in the background. \\
VIS-C3 & Motion and camera & 7 & Shake or wobble (\emph{fatal over most of the film}); objects moving by themselves or sliding; deformation or clipping during an action (\emph{fatal if the action becomes unreadable}); jumps without a cut; stiff, stuttering or frozen motion; purposeless actions; unnaturally slow, floating or near-static motion. \\
VIS-C4 & Cross-shot consistency & 7 & The protagonist becomes another person (\emph{fatal}); slight differences in the protagonist's face; unexplained changes of costume or hair, of props (\emph{fatal more than three times}) or of the set; people appearing or vanishing; confused relations between characters. \\
\midrule
AUD-C1 & Speech intelligibility & 5 & Unintelligible main lines (\emph{fatal for 30\% of the lines}); mispronunciation; swallowed or cut-off words; distorted voice; shouting that hides the words. \\
AUD-C2 & Voice naturalness and fit & 7 & Lines given to the wrong character or out of order (\emph{fatal}); a voice that contradicts the character's sex, age or role (\emph{fatal}); flat synthetic delivery; a voice that changes for the same character; emotion opposite to the scene; unnatural pauses; accent or language that contradicts the setting. \\
AUD-C3 & Background audio & 6 & Meaningless background speech (\emph{fatal twice or for 3\,s}); hiss or hum; pops or clipping; sudden breaks; reversed or weird processed voices not asked for (\emph{fatal over 30\% of the film}); audible loop seams. \\
AUD-C4 & Mix balance & 3 & Music or effects masking speech (\emph{fatal if a key line is lost}); large volume differences between voices; sudden loud effects. \\
\midrule
AV-C1 & Lip sync & 5 & A visible speaker whose mouth does not move (\emph{fatal for 30\% of the speech}); the wrong person moving the lips (\emph{fatal}); moving lips without speech; mouth movement out of rhythm; a corrupted mouth. \\
AV-C2 & Sound-event sync & 5 & Effects without a visible event (\emph{fatal twice}); key events without sound; effects early or late; effects of the wrong kind; ambience that does not fit the scene. \\
AV-C3 & Voice-over and music fit & 6 & Narration about something other than the picture (\emph{fatal over 30\% of the film}); subtitles that differ in meaning from the speech; music opposite to the mood; music that does not fit the subject or brand; sound that does not follow a change of scene; music cut abruptly at edits. \\
\bottomrule
\end{tabularx}
\end{table*}

\begin{tcolorbox}[sbprompt,title={Judge prompt of the picture call (translated, abridged)}]
\textbf{System.} You are a video reviewer. Observe only the real input media; do not take the brief, the subtitles or the picture as evidence of what is heard. Follow the program facts and the millisecond time format. Return JSON only.

\textbf{User.} You are a professional reviewer of finished short videos. Your task is not to give a score but to check, item by item, whether each check is present and to give verifiable evidence. The final score is computed by a program from your answers; you need not and should not consider what score the film will get.

Brief (not an observation; use it only to decide that an effect was requested and is therefore not a problem): \{brief\}. Genre: \{genre\}. Program facts (authoritative; do not estimate duration, aspect ratio or sound yourself): \{facts\}.
All times are whole milliseconds of the original film, $0 \le$ start $\le$ end $\le$ \{duration\}.

\pl{Each check takes one status. \emph{present}: list every instance with its time span and what happens, separately for each occurrence. \emph{absent}: state what you checked. \emph{na}: only if the object does not exist and the brief does not ask for it, quoting the brief; a requirement that is not met is present, not na. \emph{unsure}: only if it truly cannot be judged; uncertainty about severity is not a reason, list it as present.}
\pl{Rather list a defect than excuse it. ``Common in AI video'', ``possibly intended stylization'' (unless the brief asks for that style) and ``still understandable overall'' are not reasons to leave a defect out; the program weighs such factors.}
\pl{Not problems: normal cuts, transitions, fades and freeze frames; costume changes, transformations, time jumps and style changes asked for by the brief; the resolution itself; personal taste.}
\pl{Record each phenomenon only under the most fitting check.}
\pl{Program candidates are suspicious segments found automatically and may be false or incomplete. Confirm each: yes, and add it to the instances of its check; no, with a reason (\eg a normal cut, a requested effect, a face in profile); or unsure. Still check by yourself for problems the program missed.}

Review the \emph{picture}: clarity and composition, body and object structure, motion and camera, subject and scene consistency. Picture only, ignore the sound. Attached: 12 high-resolution frames with their times and face crops. Program measurements: \{summary\}. Candidates: \{candidates\}.
The subject is the main person or object of the shot; mark for every instance whether it lies on the subject. Consistency compares the whole film; changes stated by the brief or the story are not problems.

\emph{[the 29 picture checks of \cref{tab:metrics}, \eg] V4-1 The protagonist becomes another person: face, age, sex or build clearly change, so that it is obviously not the same person; compare frames at different times and the face crops. Slight differences in facial detail belong to V4-7.}

Return JSON: \texttt{\{"candidates": [\{"id", "confirm": "yes\textbar{}no\textbar{}unsure", "check\_id", "reason"\}], "checks": \{"V1-1": \{"status": "present\textbar{}absent\textbar{}na\textbar{}unsure", "instances": [\{"start\_ms", "end\_ms", "on\_subject", "note"\}], "checked", "na\_basis"\}, \dots\}\}}; checks must contain every check above.
\end{tcolorbox}

\section{Human Evaluation}
\label{sec:supp_human}
\noindent\textbf{Raters and protocol.}
Nine trained raters in three fixed groups of three rated the films on a review platform; each group took a set of tasks (49, 47, 20 of the tasks used here), and each rater independently rated every film of a task without seeing the model names.
They used the same 13 metrics with written level descriptions and reference cases, \eg for lip sync, 3: mouth movements and the onsets and offsets of speech match naturally; 2: they match broadly with local offsets; 1: clearly misaligned so that viewing suffers.
Requirement completion was judged item by item on the checklist, with 1 for met, 0.5 for partly met and 0 for not met.
Some early ratings used a five-point scale (30\% of the rating rows), which we map to three levels as 5 and 4 to 3, 3 to 2, and 2 and 1 to 1.
Each rater's overall score uses the aggregation of \cref{sec:supp_metrics}, and a film's human score is the mean of its three raters; a task without a film scores 0 for both the raters and the judge (41 cells).

\noindent\textbf{Coverage.}
We use the 116 tasks whose films were all rated by all three raters, 771 films and 2{,}313 ratings: 20 Documentary, 19 Commercial, 19 Creator, 19 Dialogue, 14 Music, 12 Product, 9 Narrative and 4 Education.
The thresholds of \cref{tab:cuts} were set on 27 development tasks, drawn by genre with a fixed seed from the first 54 completed tasks; these 27 tasks are part of the 116, so the agreement below is not measured on a fully held-out set.

\noindent\textbf{Agreement between raters.}
Within a task, a rater and the mean of the other two order the films with a mean Pearson $r$ of 0.692, and across all rated films the correlation is 0.896.
A pair of films is ordered the same way in 73.7\% of the comparisons when ties are excluded and 66.4\% when a tie counts as an error, with 10.0\% ties.
\Cref{tab:kappa} gives the agreement on single metrics: raters agree well on sound and much less on picture, expression and lip sync.

\begin{table}[h]
\centering
\caption{\textbf{Agreement on single metrics}, quadratic-weighted $\kappa$: the judge against each rater (mean over raters) and between two raters, on the delivered films of the 116 rated tasks.}
\label{tab:kappa}
\footnotesize
\setlength{\tabcolsep}{2pt}
\begin{tabular*}{\linewidth}{@{\extracolsep{\fill}}lrr@{}}
\toprule
Metric & Judge & Rater--rater \\
\midrule
Speech intelligibility & 0.52 & 0.55 \\
Voice naturalness & 0.22 & 0.37 \\
Background audio & 0.48 & 0.48 \\
Mix balance & 0.35 & 0.39 \\
Expression & 0.24 & 0.25 \\
Clarity and composition & 0.27 & 0.23 \\
Body and object structure & 0.22 & 0.26 \\
Motion and camera & 0.14 & 0.37 \\
Cross-shot consistency & 0.14 & 0.21 \\
Lip sync & 0.10 & 0.24 \\
Sound-event sync & 0.21 & 0.28 \\
Voice-over and music fit & 0.26 & 0.33 \\
\bottomrule
\end{tabular*}
\end{table}

\begin{figure}[h]
\centering
\includegraphics[width=\linewidth]{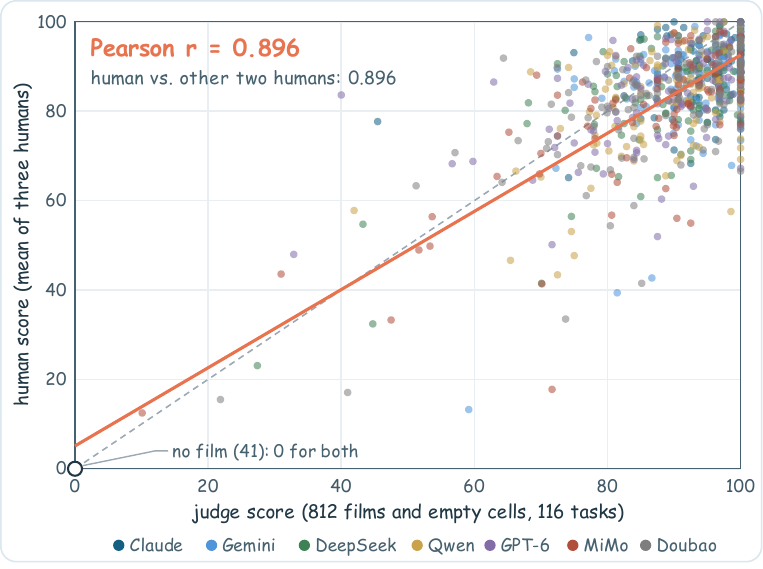}
\caption{\textbf{Judge against the mean of three humans, film level.} Each point is one film or one empty cell, scored 0 on both sides; 812 points over 116 tasks. The dashed line is $y=x$, the solid line the fit ($r=0.896$); one human against the mean of the other two reaches $r=0.896$ as well.}
\label{fig:pearson_film}
\end{figure}

\begin{figure}[h]
\centering
\includegraphics[width=\linewidth]{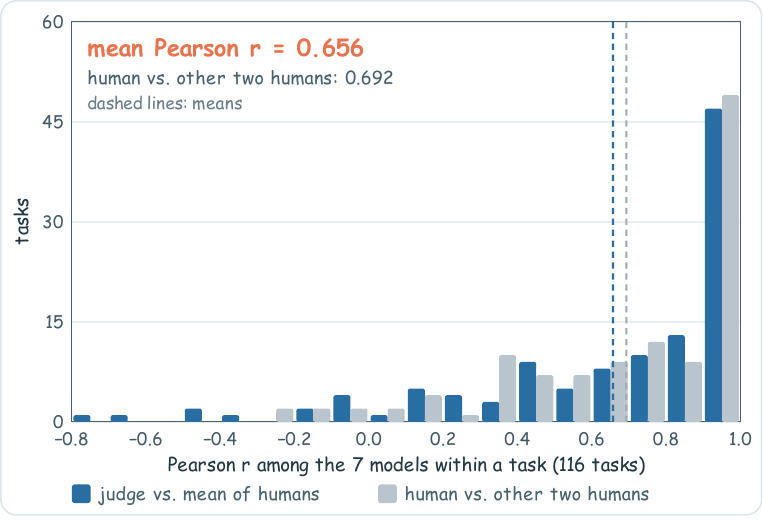}
\caption{\textbf{Within-task agreement.} For each of the 116 tasks, the Pearson $r$ between the judge and the mean of three humans (blue) and between one human and the mean of the other two (gray), computed over the films of that task. The dashed lines mark the means, 0.656 and 0.692.}
\label{fig:pearson_task}
\end{figure}

\noindent\textbf{Agreement between judge and raters.}
\Cref{tab:agree} compares the judge with the mean of the three raters.
Model-level agreement averages each model over the tasks, with 0 for a task without a film; within-task agreement compares films of the same task and averages over tasks.
Across the 116 tasks the judge correlates with the human mean at Pearson $r=0.997$ at the model level and 0.656 within a task, against 0.692 between a rater and the mean of the other two, and it orders 72.3\% of film pairs within a task as the raters do (63.5\% when a tie on either side counts as an error).
\Cref{tab:human_rank} lists both rankings; the judge scores 5.6 points higher on average, with a mean absolute difference of 8.6 points, but orders all 21 model pairs identically. \Cref{fig:pearson_film,fig:pearson_task} break the correlation down: at the film level the judge agrees with the human mean as closely as a human does, while within a task most scores cluster near 0.9 with a left tail of hard tasks that pulls the mean down.
The agreement supports the scoring protocol and the ranking on these tasks; with nine Narrative tasks among them, it does not by itself validate the ranking on all 160 tasks, where DeepSeek V4.1 Flash moves ahead of Doubao Seed 2.1 Pro.

\begin{table}[h]
\centering
\caption{\textbf{Judge against the mean of three raters.} Pairs: model pairs ordered the same way; $r$: Pearson correlation. Within a task, accuracy is the share of film pairs ordered as by the raters, excluding ties or counting a tie by the judge as an error.}
\label{tab:agree}
\footnotesize
\setlength{\tabcolsep}{1.5pt}
\begin{tabular*}{\linewidth}{@{\extracolsep{\fill}}lrrrrrr@{}}
\toprule
 & \multicolumn{2}{c}{Models} & \multicolumn{4}{c}{Within a task} \\
\cmidrule(lr){2-3}\cmidrule(l){4-7}
Tasks & Pairs & $r$ & $r$ & Acc. & Acc.\,tie & Ties \\
\midrule
All (116) & 21/21 & 0.997 & 0.656 & 0.723 & 0.635 & 12.2\% \\
\midrule
Rater--rater (116) & -- & -- & 0.692 & 0.737 & 0.664 & 10.0\% \\
\bottomrule
\end{tabular*}
\end{table}

\begin{table}[h]
\centering
\caption{\textbf{Rankings on the 116 human-rated tasks} (rank in parentheses).}
\label{tab:human_rank}
\footnotesize
\begin{tabular*}{\linewidth}{@{\extracolsep{\fill}}lrr@{}}
\toprule
Model & Raters & Judge \\
\midrule
Claude Opus 5.5 & 88.0 (1) & 92.8 (1) \\
Gemini 4 Argon & 84.9 (2) & 91.4 (2) \\
GPT-6 Astra & 83.6 (3) & 88.3 (3) \\
DeepSeek V4.1 Flash & 81.8 (4) & 87.6 (4) \\
Doubao Seed 2.1 Pro & 81.3 (5) & 87.2 (5) \\
Qwen3.8 Max & 77.0 (6) & 83.3 (6) \\
MiMo v2.6 Pro & 60.1 (7) & 64.9 (7) \\
\bottomrule
\end{tabular*}
\end{table}

\section{Robustness of the Ranking}
\label{sec:supp_schemes}
\Cref{tab:schemes} scores every model over all 160 tasks, with 0 for a task without a film, under variants of the protocol.
\emph{Continuous} replaces the three levels by $1 - \min(c, b + 1.5)/(b + 1.5)$ for a cost $c$, capped at 0.33 when a fatal check is present; it almost never ties but follows the human ranking slightly less closely than the levels.
\emph{Pass 1} and \emph{Pass 2} use one pass of the judge each, and \emph{Equal} weights the four dimensions equally in every genre.
All five variants give the same ranking.
Within a single film, the overall scores of the two passes differ by 4.5 points on average, and 87.0\% of the metric levels agree, so we report the mean of two passes.

\begin{table}[h]
\centering
\caption{\textbf{The ranking under variants of the protocol}, mean overall score over all 160 tasks with 0 for a task without a film. Best in \textbf{bold}.}
\label{tab:schemes}
\footnotesize
\setlength{\tabcolsep}{3pt}
\begin{tabular*}{\linewidth}{@{\extracolsep{\fill}}lrrrrrr@{}}
\toprule
Model & Done & Main & Cont. & Pass 1 & Pass 2 & Equal \\
\midrule
Claude Opus 5.5 & 156 & \textbf{92.2} & \textbf{93.1} & \textbf{92.9} & \textbf{93.2} & \textbf{91.8} \\
Gemini 4 Argon & 159 & 92.0 & 93.0 & 92.8 & 93.1 & 91.4 \\
GPT-6 Astra & 160 & 88.8 & 90.9 & 89.9 & 90.5 & 87.8 \\
DeepSeek V4.1 Flash & 156 & 88.3 & 89.9 & 89.7 & 89.5 & 87.5 \\
Doubao Seed 2.1 Pro & 160 & 86.9 & 89.5 & 88.6 & 88.6 & 85.9 \\
Qwen3.8 Max & 152 & 83.9 & 86.0 & 85.6 & 85.4 & 83.0 \\
MiMo v2.6 Pro & 120 & 63.4 & 65.2 & 64.8 & 64.2 & 62.8 \\
\bottomrule
\end{tabular*}
\end{table}

\section{Results by Genre, Task Type and Sound Label}
\label{sec:supp_breakdown}
\Cref{tab:breakdown} gives the mean overall score of every model by genre, task type and sound label, with 0 for each task on which a model delivered no film.
Claude Opus 5.5 is best in Narrative, Commercial, Creator and Dialogue, Gemini 4 Argon in Product, Education and Documentary, and DeepSeek V4.1 Flash and GPT-6 Astra tie at 88.6 in Music.
Dialogue and Music are the hardest genres, with means of 81.7 and 82.1, and Documentary and Product the easiest, at 89.0 and 88.0.
MiMo v2.6 Pro is lowest in most genres; its gap is largely explained by its 40 tasks without a film, seven each in Narrative, Commercial and Dialogue.

\begin{table*}[t]
\centering
\caption{\textbf{Mean overall score by genre, task type and sound label} (0--100). $n$: tasks; each model is scored on all $n$ tasks, with 0 for a task without a film. Best per row in \textbf{bold}; All is the mean over the seven models.}
\label{tab:breakdown}
\small
\setlength{\tabcolsep}{3pt}
\begin{tabular*}{\linewidth}{@{\extracolsep{\fill}}lrrrrrrrrr@{}}
\toprule
 & $n$ & Claude & Gemini & DeepSeek & Qwen & GPT-6 & MiMo & Doubao & All \\
\midrule
\multicolumn{10}{@{}l}{\emph{Genre}} \\
Narrative & 20 & \textbf{95.3} & 90.7 & 91.1 & 80.5 & 83.8 & 57.1 & 81.0 & 82.8 \\
Commercial & 20 & \textbf{92.6} & 91.6 & 88.4 & 82.8 & 86.2 & 56.1 & 90.7 & 84.1 \\
Product & 20 & 86.8 & \textbf{97.8} & 95.5 & 91.3 & 93.3 & 62.2 & 88.6 & 88.0 \\
Education & 20 & 90.2 & \textbf{94.7} & 87.7 & 85.2 & 93.1 & 64.8 & 88.3 & 86.3 \\
Music & 20 & 87.0 & 85.4 & \textbf{88.6} & 81.7 & 88.6 & 61.1 & 82.0 & 82.1 \\
Documentary & 20 & 93.5 & \textbf{93.6} & 88.0 & 84.4 & 90.8 & 79.9 & 92.7 & 89.0 \\
Creator & 20 & \textbf{96.6} & 93.2 & 86.8 & 84.1 & 84.6 & 73.5 & 88.3 & 86.7 \\
Dialogue & 20 & \textbf{95.8} & 89.0 & 80.7 & 81.3 & 89.6 & 52.1 & 83.3 & 81.7 \\
\midrule
\multicolumn{10}{@{}l}{\emph{Task type}} \\
Open Brief & 32 & \textbf{95.2} & 93.6 & 91.0 & 85.2 & 92.1 & 69.2 & 89.1 & 87.9 \\
Constraint & 59 & 89.8 & \textbf{91.0} & 86.4 & 85.3 & 87.6 & 59.0 & 85.5 & 83.5 \\
Asset-cond. & 34 & 92.7 & \textbf{94.4} & 90.4 & 84.2 & 91.2 & 61.7 & 86.7 & 85.9 \\
Reference-cond. & 15 & \textbf{94.9} & 89.1 & 87.5 & 85.3 & 77.4 & 56.3 & 84.0 & 82.1 \\
Audio-led & 20 & \textbf{91.8} & 90.6 & 86.9 & 76.3 & 91.3 & 74.7 & 89.8 & 85.9 \\
\midrule
\multicolumn{10}{@{}l}{\emph{Sound label (multi-label)}} \\
Speech & 89 & \textbf{94.3} & 93.2 & 89.8 & 83.7 & 89.7 & 65.1 & 88.4 & 86.3 \\
Dialogue & 55 & \textbf{94.4} & 89.2 & 86.6 & 82.5 & 85.2 & 60.0 & 83.1 & 83.0 \\
Music & 157 & \textbf{92.2} & 92.0 & 88.7 & 84.0 & 88.7 & 64.0 & 87.3 & 85.3 \\
Sound effects & 128 & \textbf{92.1} & 92.0 & 88.5 & 83.4 & 88.0 & 62.9 & 86.4 & 84.8 \\
\bottomrule
\end{tabular*}
\end{table*}

\section{Environment Errors}
\label{sec:supp_infra}
\paragraph{Runs without a film.}
Of 1{,}120 runs, 57 delivered no film.
Twelve were interrupted by the model API (Qwen3.8 Max 6, Claude Opus 5.5 4, DeepSeek V4.1 Flash 2) and one DeepSeek V4.1 Flash run reached the time limit.
In 40 runs the model ended its session normally without calling \texttt{export} (MiMo v2.6 Pro 36, Qwen3.8 Max 2, DeepSeek V4.1 Flash 1, Gemini 4 Argon 1), and 4 further MiMo v2.6 Pro runs exited with an error.
In a typical MiMo v2.6 Pro case the model announces the final mix in its last message and stops before exporting.
All 57 runs score 0 in the main results.
\Cref{tab:zero} also scores each model without the 13 runs that the API or the time limit cut short, since these are not the model's choice.

\begin{table}[h]
\centering
\caption{\textbf{Mean overall score under three treatments of environment problems.} Main: all 160 tasks, 0 for a task without a film. No API: the 13 runs interrupted by the API or the time limit leave the denominator. No service: the 150 delivered runs with five or more service-side tool errors leave the denominator.}
\label{tab:zero}
\small
\begin{tabular*}{\linewidth}{@{\extracolsep{\fill}}lrrr@{}}
\toprule
Model & Main & No API & No service \\
\midrule
Claude Opus 5.5 & 92.2 & 94.6 & 91.9 \\
Gemini 4 Argon & 92.0 & 92.0 & 92.5 \\
GPT-6 Astra & 88.8 & 88.8 & 89.1 \\
DeepSeek V4.1 Flash & 88.3 & 90.0 & 88.2 \\
Doubao Seed 2.1 Pro & 86.9 & 86.9 & 87.0 \\
Qwen3.8 Max & 83.9 & 87.2 & 83.5 \\
MiMo v2.6 Pro & 63.4 & 63.4 & 62.0 \\
\bottomrule
\end{tabular*}
\end{table}

\paragraph{Service-side tool errors.}
Tool services occasionally failed independently of the model, most often image editing, lip sync and speech recognition, for example when a GPU worker crashed or a request reached a worker serving another tool.
The share of delivered runs with at least one such error ranges from 14\% (Doubao Seed 2.1 Pro) to 45\% (Claude Opus 5.5), partly because models that call more tools meet more failures.
Removing the 150 delivered runs with five or more service errors changes each model's score by at most 1.3 points and only swaps Claude Opus 5.5 and Gemini 4 Argon, which are 0.2 points apart in the main results; the affected and the other delivered runs both average 89.6.

\section{Objective Measurement}
\label{sec:supp_measure}
\begin{table}[h]
\centering
\caption{\textbf{Objective defects of delivered films.} Share of films (\%) that are silent (integrated loudness below $-60$\,LUFS), contain more than 10\,s of silence, more than 2\,s of black frames or more than 10\,s of frozen frames, and washed-out frames in at least 40\% of sampled frames, and the cut rate relative to the human source video (paired ratio). Loudness follows ITU-R BS.1770 \cite{itu2023bs1770}.}
\label{tab:defects}
\small
\setlength{\tabcolsep}{3pt}
\resizebox{\linewidth}{!}{%
\begin{tabular}{@{}lrrrrrr@{}}
\toprule
Model & Silent & Sil.\,$>$10s & Black & Frozen & Washed & Cuts/human \\
\midrule
Claude Opus 5.5 & 0.0 & 0.6 & 9.6 & 5.8 & 2.6 & 0.46 \\
Gemini 4 Argon & 0.0 & 1.3 & 1.3 & 10.1 & 0.6 & 0.29 \\
DeepSeek V4.1 Flash & 1.9 & 3.2 & 18.6 & 16.7 & 3.3 & 0.46 \\
Qwen3.8 Max & 0.0 & 5.9 & 19.1 & 19.1 & 2.6 & 0.29 \\
GPT-6 Astra & 0.0 & 2.5 & 2.5 & 14.4 & 4.4 & 0.35 \\
MiMo v2.6 Pro & \textbf{4.2} & 9.2 & 4.2 & 25.0 & 1.7 & 0.21 \\
Doubao Seed 2.1 Pro & 0.0 & 8.8 & 14.4 & 20.6 & 3.8 & 0.35 \\
\bottomrule
\end{tabular}}
\end{table}

\cref{tab:defects} gives the defect rates per model. All measurements use FFmpeg filters on the delivered file.
Integrated loudness, loudness range and true peak come from the \texttt{ebur128} filter, which implements ITU-R BS.1770 \cite{itu2023bs1770,ebu2023r128}; a film is silent when its integrated loudness is below $-60$\,LUFS.
Silence uses \texttt{silencedetect} at $-45$\,dB for at least 1.5\,s, black frames use \texttt{blackdetect} (pixel threshold 0.10, at least 0.5\,s), and frozen frames use \texttt{freezedetect} (noise 0.003, at least 2\,s).
Cuts are frames whose scene-change score exceeds 0.3 after scaling to 320 pixels width.
A film is washed out when at least 40\% of 16 sampled frames have mean luma above 185 and contrast below 35.
The human source video of each task was measured with the same pipeline: it cuts 12.0 times per minute at the median, with a median shot of 4.1\,s, against 1.3--7.3 cuts per minute and median shots of 6.0--30.4\,s for the models.
The paired ratio of model to human cut rate, with a task-level bootstrap 95\% CI, is 0.46 for both Claude Opus 5.5 [0.37, 0.56] and DeepSeek V4.1 Flash [0.39, 0.55], and 0.21 [0.16, 0.26] for MiMo v2.6 Pro at the low end; every interval excludes 1.

\section{Automatic Trace Analysis}
\label{sec:supp_trace}
\noindent\textbf{Workflow graphs.}
The environment logs, for every call, the files it read and wrote and the call that produced each file it read.
Each call is a node, and an edge links the producer of a file to every later call that read it.
The final \texttt{export} is the root of the film, and its ancestors are the calls whose output reaches the film.
Production nodes are successful calls other than \texttt{export} and the checks (\texttt{sample\_frames}, \texttt{inspect}, \texttt{asr}, \texttt{scene\_detect}, \texttt{beat\_detect}, \texttt{search} and \texttt{fetch\_url}).
In \cref{tab:process}, Used is the share of production nodes among the ancestors of the export, and Depth is the number of nodes on the longest path that ends at the export.
The logs yield 1{,}062 graphs.

\noindent\textbf{Production steps.}
\Cref{fig:workflow} counts a step in a delivered run by the following rules.
\emph{Script first}: more than 300 characters of text or reasoning before the first tool call.
\emph{Keyframe}: the first image call (\texttt{t2i} or \texttt{image\_edit}) comes before the first video generation.
\emph{Character lock}: at least one \texttt{image\_edit} call.
\emph{Review generation}: a \texttt{sample\_frames} call after the first video generation.
\emph{Trim}: at least one \texttt{trim} call.
\emph{Mix}: the first \texttt{mix\_audio} or \texttt{audio\_mix} call comes before the first export.
\emph{Inspect}: an \texttt{inspect} call before the first export.
\emph{Re-export}: more than one \texttt{export} call.

\noindent\textbf{Error-mode proxies.}
The proxies of \cref{tab:errors} are computed on the 1{,}063 delivered films.
\emph{Character drift}: cross-shot consistency below the top level, among films where it applies.
\emph{AV misalignment}: lip sync or sound-event sync below the top level, whichever is lower, among films where either applies; a silent film takes the lowest level on both by rule.
\emph{Silent damage}: an integrated loudness below $-60$\,LUFS, or washed-out frames in at least 40\% of 16 sampled frames (\cref{sec:supp_measure}).
\emph{Tool-length pacing}: among shots longer than 0.3\,s, at least half last within 0.3\,s of 5\,s or 8\,s, the segment lengths of the generator.

\noindent\textbf{Pair analysis.}
For each task we take the highest- and the lowest-scoring delivered film, A and B, and keep the 142 tasks on which they differ by at least 10 points (median gap 20).
Gemini 3.8 Flash at temperature 0 receives, as text and without model names, the brief; for each film its overall, dimension and metric scores, every defect the judge found in either pass with its time span and note, and the program measurements; and the full workflow graph, one line per call with its tool, a summary of its arguments and result, the calls whose outputs it read, whether its output reaches the film, and the model's own words after every check.
It returns up to three quality losses of B, each with the defect and the judge checks it rests on, the calls of B that introduced it, the checks that followed and what came of them, one of five abilities, and what A did differently.
A program verifies that every cited call belongs to B and every cited check appears among B's defects; an invalid answer is asked again once, which was never needed.
\Cref{tab:pairs} gives the frequencies.
Two automatic checks support the attributions: 99.7\% of the calls named as the origin of a loss are production calls and 99.0\% of the calls named as checks are checking tools, and in four of the six pairs whose weaker film is silent, where the cause is known, the analysis traces the loss to the mixing call and classifies it as unverified.
We also checked 79 of these losses by hand, in 27 pairs whose weaker film has sound, against frames and audio of the film at the flagged time spans and the logged calls and checks; each loss was judged correct, partly correct or wrong, and all 79 were judged correct.

\begin{tcolorbox}[sbprompt,title={Pair analysis prompt (translated from Chinese, abridged)}]
You are an analyst of video production workflows. On the same task, two AI directors each made a film with the same atomic tools: A scored high and B scored low. You receive the judge's findings for both films and their complete workflow graphs, including every tool call, its outputs and results, the calls whose outputs it read, and what the director wrote after each check. You cannot see the videos; infer only from these records and do not invent facts that are not in them.

Find at most three quality losses where B lost the most against A. For each give:
\pl{\emph{defect} and \emph{metric}, citing the judge's findings, and \emph{evidence}, the judge check IDs, which must appear among B's findings;}
\pl{\emph{origin\_calls}, the calls of B that introduced the problem; if it comes from a missing step, the call after which that step should have been added;}
\pl{\emph{detection}: \emph{not\_checked}, no later check could have found it; \emph{checked\_missed}, a relevant check followed but the director's words or actions show it was not noticed; \emph{noticed\_not\_fixed}; or \emph{partly\_fixed}; with the \emph{check\_calls};}
\pl{\emph{capability}: \emph{state}, a character, scene, voice or sync drifts across generation and editing steps; \emph{verify}, an output has a perceptual defect but no tool reported an error, including wrong parameters whose result looks fine to the tool, and the director did not verify it; \emph{effect}, a decision follows tool defaults rather than the effect on the film, such as cutting at the tool's clip length, making no alternatives or mixing without checking the result; \emph{execution}, a tool reported an error or failed and the director did not recover; \emph{brief}, the brief was misread or a required item was left out;}
\pl{\emph{contrast}, what A did differently at that point, citing A's calls, and a \emph{confidence}.}
Finally give the \emph{main\_capability} of B and a one-sentence summary. Return JSON only.
\end{tcolorbox}

\begin{table}[h]
\centering
\caption{\textbf{Pair analysis of the weaker film on 142 tasks.} Share (\%) of the quality losses by ability and by what later checks did, over all pairs and over the pairs whose weaker film has sound.}
\label{tab:pairs}
\footnotesize
\begin{tabular*}{\linewidth}{@{\extracolsep{\fill}}lrr@{}}
\toprule
 & All & Film B has sound \\
\midrule
Pairs / losses & 142 / 415 & 136 / 401 \\
\midrule
Unverified defect & 65.5 & 65.3 \\
State drift & 15.4 & 15.7 \\
Tool default over effect & 16.6 & 16.5 \\
Unrecovered tool error & 1.4 & 1.5 \\
Misread brief & 1.0 & 1.0 \\
\midrule
Checked but missed & 57.8 & 57.1 \\
Never checked & 33.5 & 34.4 \\
Noticed, not fixed & 7.2 & 7.2 \\
Partly fixed & 1.4 & 1.2 \\
\bottomrule
\end{tabular*}
\end{table}

\begin{figure*}[p]
\centering
\includegraphics[width=\linewidth]{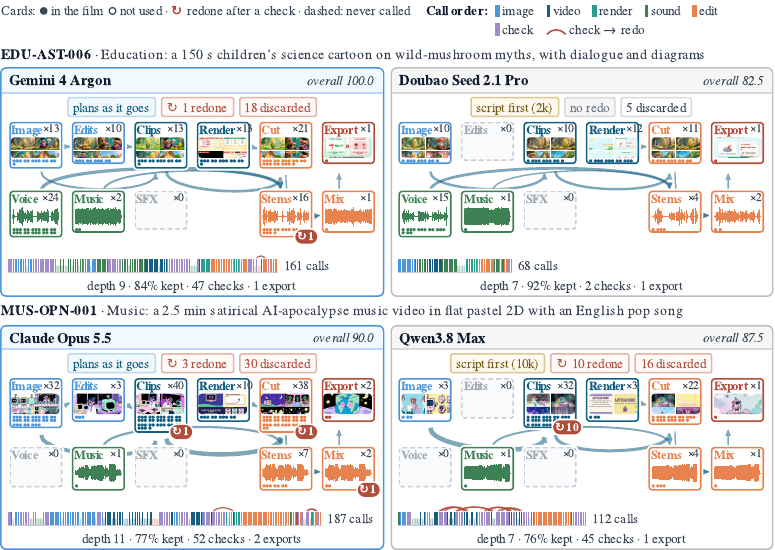}
\caption{\textbf{Workflow graphs on an Education and a Music task}, drawn as in \cref{fig:workflow_case}.}
\label{fig:wg_app1}
\end{figure*}

\begin{figure*}[p]
\centering
\includegraphics[width=\linewidth]{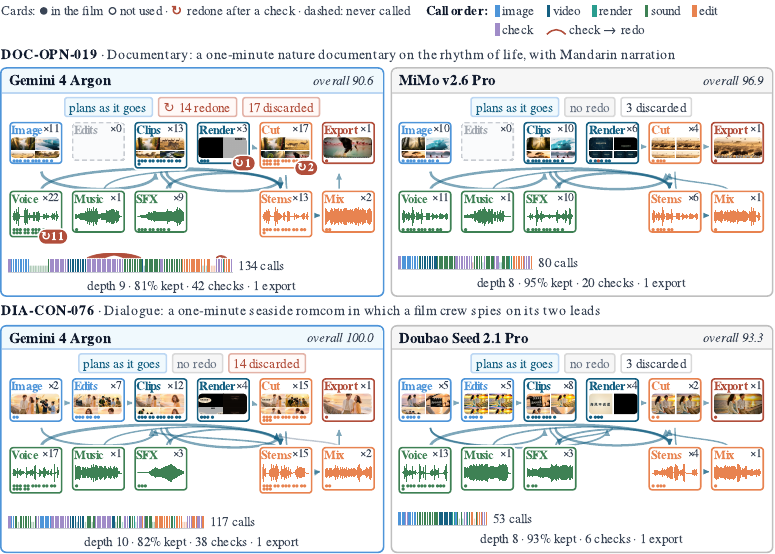}
\caption{\textbf{Workflow graphs on a Documentary and a Dialogue task}, drawn as in \cref{fig:workflow_case}.}
\label{fig:wg_app2}
\end{figure*}

\noindent\textbf{More workflow graphs.}
\Cref{fig:wg_app1,fig:wg_app2} extend \cref{fig:workflow_case} to four more genres, one task per genre, each with one of the two top-ranked models and a lower-ranked model on the same task.
For each genre we considered the 16:9 tasks on which both models delivered a film and neither run had more than five tool errors, and took a task whose two graphs are close to the median graph of their models in number of nodes, depth, share of production nodes that reach the film, redone and discarded takes and planning text; every chosen task is among the five closest of its genre.
The pairs illustrate how the models orchestrate production, not which film is better: on two of the six tasks, NAR-AST-004 and DOC-OPN-019, the lower-ranked model's film scores higher.
Cards group the calls of a run by tool and show their real outputs; a filled dot is a call whose output reaches the film.
A product counts as redone after a check when a check reads it or its descendants, the product does not reach the film, and a later call of the same tool writes a new version of the same file, identified by its name up to a version suffix such as \texttt{\_v2} or \texttt{b}, that does.
Takes discarded are successful production calls that do not reach the film, and the strip under each graph lists the calls in order, coloured by stage.
Script first means that the model wrote more than 1{,}000 characters before its first tool call.

\section{Qualitative Comparison}
\label{sec:supp_qual}
\begin{figure*}[p]
\centering
\includegraphics[width=\linewidth]{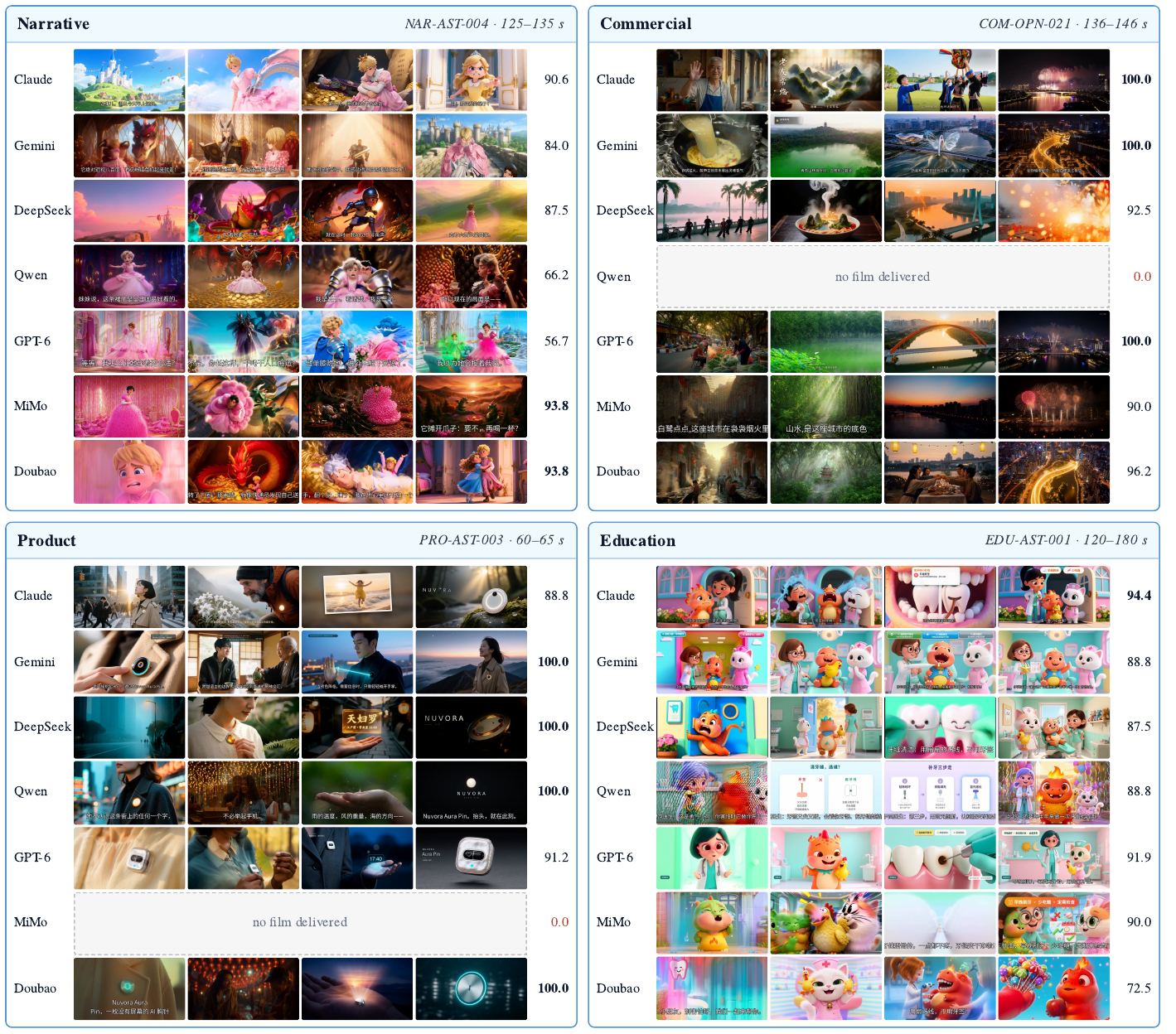}
\caption{\textbf{Qualitative comparison, Narrative to Education.} Four frames of each film at 10, 37, 63 and 90\% of its length; the number is the overall score, 0 for a task without a film. Best per task in bold.}
\label{fig:qual1}
\end{figure*}

\begin{figure*}[p]
\centering
\includegraphics[width=\linewidth]{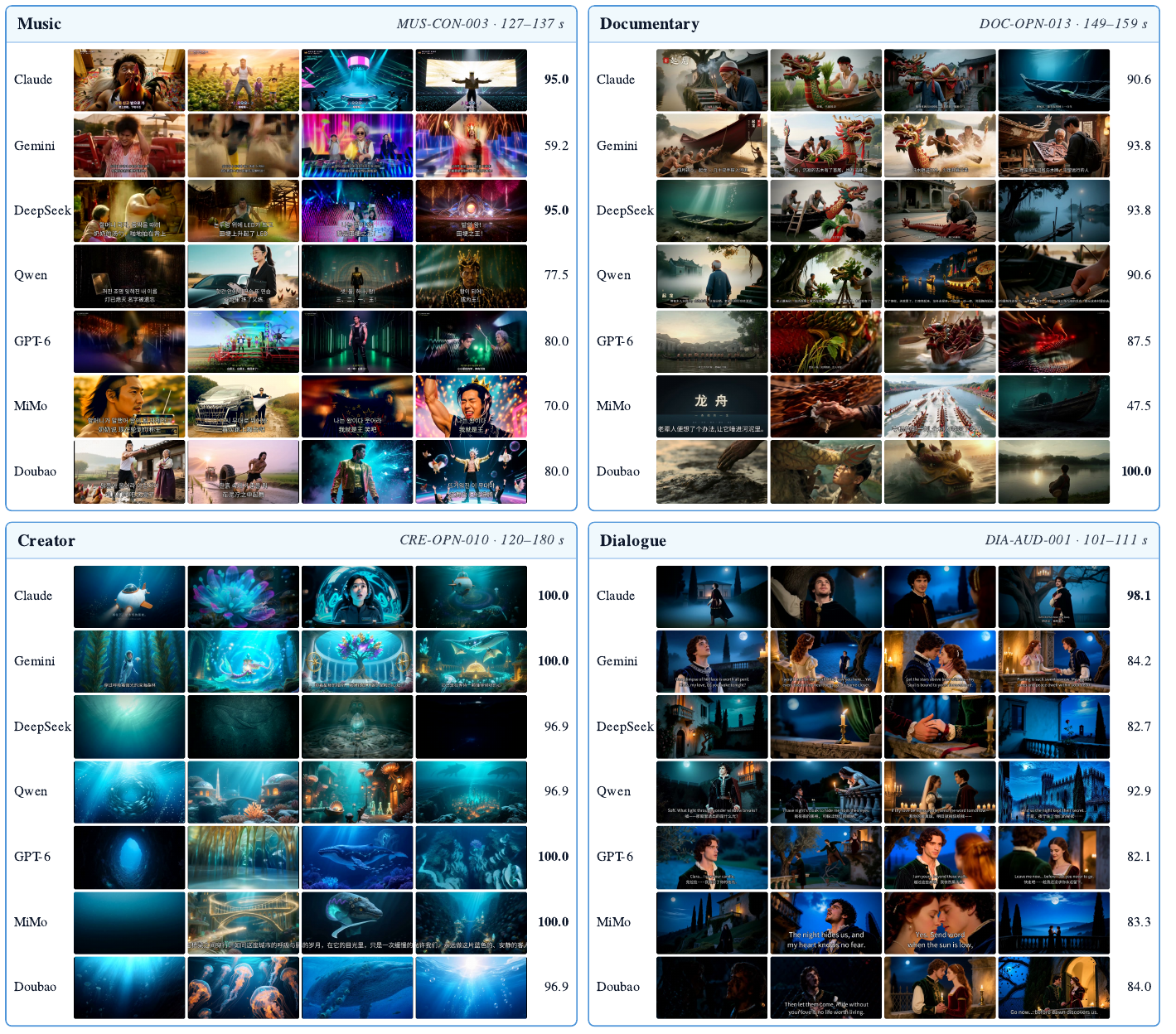}
\caption{\textbf{Qualitative comparison, Music to Dialogue.} Same layout as \cref{fig:qual1}.}
\label{fig:qual2}
\end{figure*}

For every genre we drew one task at random with a fixed seed among its 16:9 tasks, so that all frames share one aspect ratio; the draw does not look at the scores.
\Cref{tab:qual_tasks} summarizes the eight briefs, and \cref{fig:qual1,fig:qual2} show four frames of every delivered film, taken at 10, 37, 63 and 90\% of its length.
Every model is shown on every task: a model that delivered no film appears as an empty row with a score of 0, as Qwen3.8 Max on COM-OPN-021 and MiMo v2.6 Pro on PRO-AST-003.

\begin{table}[h]
\centering
\caption{\textbf{Tasks of the qualitative comparison} (briefs summarized in translation).}
\label{tab:qual_tasks}
\footnotesize
\begin{tabularx}{\linewidth}{@{}ll>{\raggedright\arraybackslash}X@{}}
\toprule
Genre & Task & Brief \\
\midrule
Narrative & NAR-AST-004 & A fantasy comedy: a prince trying on his sister's pink gown is carried off by a dragon by mistake; the dragon turns into a silver-haired nobleman and the rescuing knight mistakes him, told through the prince's quips in Chinese. \\
Commercial & COM-OPN-021 & A 140\,s tourism film for Nanning built on its local spirit of old friends: street life, nature, crafts, local food, the Yong River, ASEAN elements and a night dragon dance with fireworks. \\
Product & PRO-AST-003 & An advertisement for a fictional screenless AI pin worn in city and nature scenes, recognizing objects, translating, taking photos and projecting onto the palm. \\
Education & EDU-AST-001 & A 150\,s 3D cartoon on dental care for children: a little dinosaur afraid of the dentist is calmed and treated; it must explain flossing, the steps of a filling and daily care, with dialogue only. \\
Music & MUS-CON-003 & A 132\,s surreal K-pop music video in which a former idol is coaxed back to the stage by his grandmother, a girl and his manager, with Korean lyrics and Chinese subtitles. \\
Documentary & DOC-OPN-013 & A 154\,s film on the dragon-boat tradition, following one boat through five rituals from awakening to burial, with male narration and folk percussion. \\
Creator & CRE-OPN-010 & A 150\,s cinematic CG exploration of a calm undersea civilization with bioluminescent plants and giant sea creatures. \\
Dialogue & DIA-AUD-001 & A Renaissance balcony love scene in English dialogue: the young man climbs to the balcony, the lovers pledge themselves and part. \\
\bottomrule
\end{tabularx}
\end{table}

\end{document}